\documentclass[10pt,twocolumn,letterpaper]{article}

\usepackage[pagenumbers]{wacv} % To force page numbers, e.g. for an arXiv version

\usepackage{graphicx}
\usepackage{amsmath}
\usepackage{amssymb}
\usepackage{booktabs}

\usepackage{caption}
\usepackage{subcaption}
\usepackage{color}
\usepackage{url} 
\usepackage{amsfonts} 
\usepackage{nicefrac}   
\usepackage{multirow}
\usepackage{multicol}
\usepackage{colortbl}
\usepackage{makecell}
\usepackage{array}
\usepackage{pifont}%
\usepackage{tabularx}
\usepackage[pagebackref,breaklinks,colorlinks]{hyperref}

\usepackage[capitalize]{cleveref}
\crefname{section}{Sec.}{Secs.}
\Crefname{section}{Section}{Sections}
\Crefname{table}{Table}{Tables}
\crefname{table}{Tab.}{Tabs.}

\definecolor{lightgray}{rgb}{0.9,0.9,0.9}

\def\wacvPaperID{584} % *** Enter the WACV Paper ID here
\def\confName{WACV}
\def\confYear{2025}

\begin{document}
\newcolumntype{Y}{>{\centering\arraybackslash}X}
\newcolumntype{Z}{>{\hsize=.75\hsize\linewidth=\hsize\centering\arraybackslash}X}
\newcolumntype{V}{>{\hsize=1.25\hsize\linewidth=\hsize\centering\arraybackslash}X}
\newcolumntype{C}{>{\columncolor{lightgray}}Y}
%%%%%%%%% TITLE - PLEASE UPDATE
\title{Feature Augmentation based Test-Time Adaptation}

\author{Younggeol Cho\footnotemark[1] \qquad
Youngrae Kim\footnotemark[1] \qquad
Junho Yoon \qquad
Seunghoon Hong \qquad
Dongman Lee\\[0.5ex]
Korea Advanced Institute of Science and Technology (KAIST) \\
{\tt\small \{rangewing, youngrae.kim, vpdtlrdl, seunghoon.hong, dlee\}@kaist.ac.kr} 
% For a paper whose authors are all at the same institution,
% omit the following lines up until the closing ``}''.
% Additional authors and addresses can be added with ``\and'',
% just like the second author.
% To save space, use either the email address or home page, not both
}
\maketitle

\renewcommand{\thefootnote}{\fnsymbol{footnote}}
\footnotetext[1]{Equal contribution}
\renewcommand{\thefootnote}{\arabic{footnote}}
\setcounter{footnote}{0}
%%%%%%%%% ABSTRACT
\begin{abstract}
    Test-time adaptation (TTA) allows a model to be adapted to an unseen domain without accessing the source data. Due to the nature of practical environments, TTA has a limited amount of data for adaptation. Recent TTA methods further restrict this by filtering input data for reliability, making the effective data size even smaller and limiting adaptation potential. To address this issue, We propose Feature Augmentation based Test-time Adaptation (FATA), a simple method that fully utilizes the limited amount of input data through feature augmentation. 
    FATA employs Normalization Perturbation to augment features and adapts the model using the FATA loss, which makes the outputs of the augmented and original features similar.
    FATA is model-agnostic and can be seamlessly integrated into existing models without altering the model architecture. We demonstrate the effectiveness of FATA on various models and scenarios on ImageNet-C and Office-Home, validating its superiority in diverse real-world conditions.
    % Code is available at \url{https://anonymous.4open.science/r/FATA/}.
    %, and we plan to release the code officially upon acceptance.
\end{abstract} 
\section{Introduction}
\label{ch:intro}

Deep learning models have significantly enhanced the performance of computer vision applications~\cite{he2016deep,dosovitskiy2020image}. However, real-world scenarios often present challenges such as performance degradation caused by domain shifts between training and target domain. To mitigate this gap, unsupervised domain adaptation (UDA)~\cite{pan2010domain_uda,sun2019unsupervised,patel2015visual_uda,long2016unsupervised_uda,tsai2018learning_uda_adv,ganin2015unsupervised_uda_adv,long2015learning_uda_discrepancy_loss} and test-time training (TTT) techniques~\cite{sun2020test_ttt, liu2021ttt++} have been proposed. These methods typically rely on adapting models to unseen domains using extensive source data during testing, which is often impractical due to limited computational resources and privacy issues.
Recently, fully test-time adaptation (TTA) methods~\cite{wang2021tent,niu2022towards_SAR,chen2022contrastivetta, niu2022efficient, lee2024entropy_deyo} have emerged, enabling online adaptation of trained models to target environments without the need for source data or labels. The dominant paradigm among TTA methods involves minimizing entropy loss while updating the affine parameters of Batch Normalization~\cite{ioffe2015batchnorm} layers, as initially proposed by TENT~\cite{wang2021tent}, which demonstrates the correlation between entropy and accuracy. Extending TENT, several methods 
%such as Niu \textit{et al.}~\cite{niu2022efficient} 
propose sample selection based entropy minimization, which filters out unreliable or redundant samples. For example, Niu \textit{et al.}~\cite{niu2022efficient} demonstrates that not all data samples are reliable and performs sample selection based on entropy and sample weighting based on the reliability for each sample. Similarly, SAR~\cite{niu2022towards_SAR} filters out samples with high entropy. % and 
DeYO~\cite{lee2024entropy_deyo} also uses entropy for sample selection and filters out harmful samples that degrades the adaptation process, %where DeYO also uses 
using structure or shape in the data for further sampling.
% selects samples containing helpful structure or shape for training. 

\begin{figure*}[ht!]
    \centering
    \begin{subfigure}[b]{0.36\linewidth} %0.51
        \includegraphics[width=\linewidth]{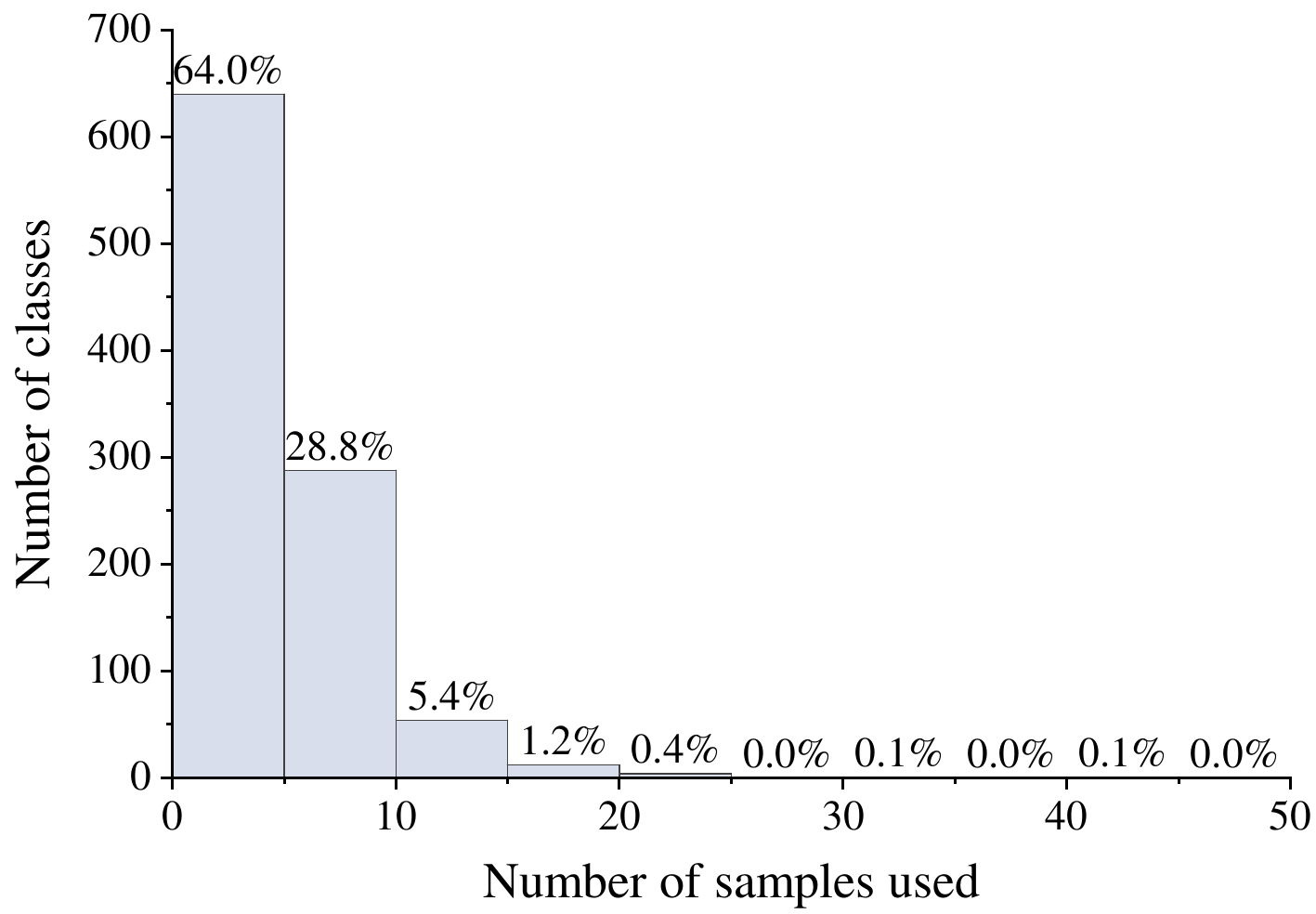}
        \caption{Count of classes for number of selected samples\\\ }
        \label{fig:intro/classes}
    \end{subfigure}
    % \hspace{1em}
    \hfill
    \begin{subfigure}[b]{0.31\linewidth} %0.44
        \includegraphics[width=\linewidth]{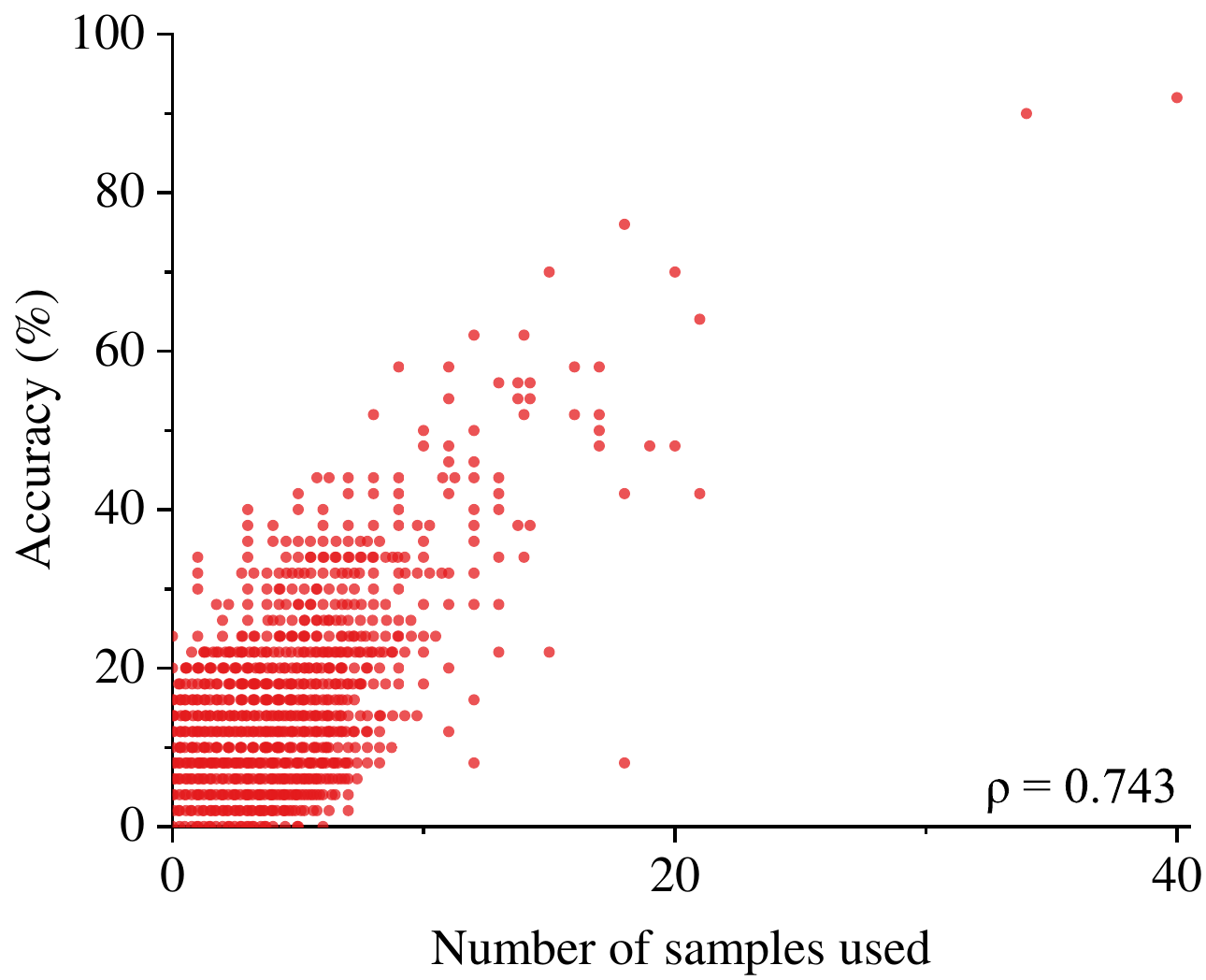}
        \caption{Accuracy of classes for number of samples used}
        \label{fig:intro/corr}
    \end{subfigure}
    \hfill
    \begin{subfigure}[b]{0.32\linewidth}
        \centering
        \includegraphics[width=\linewidth]{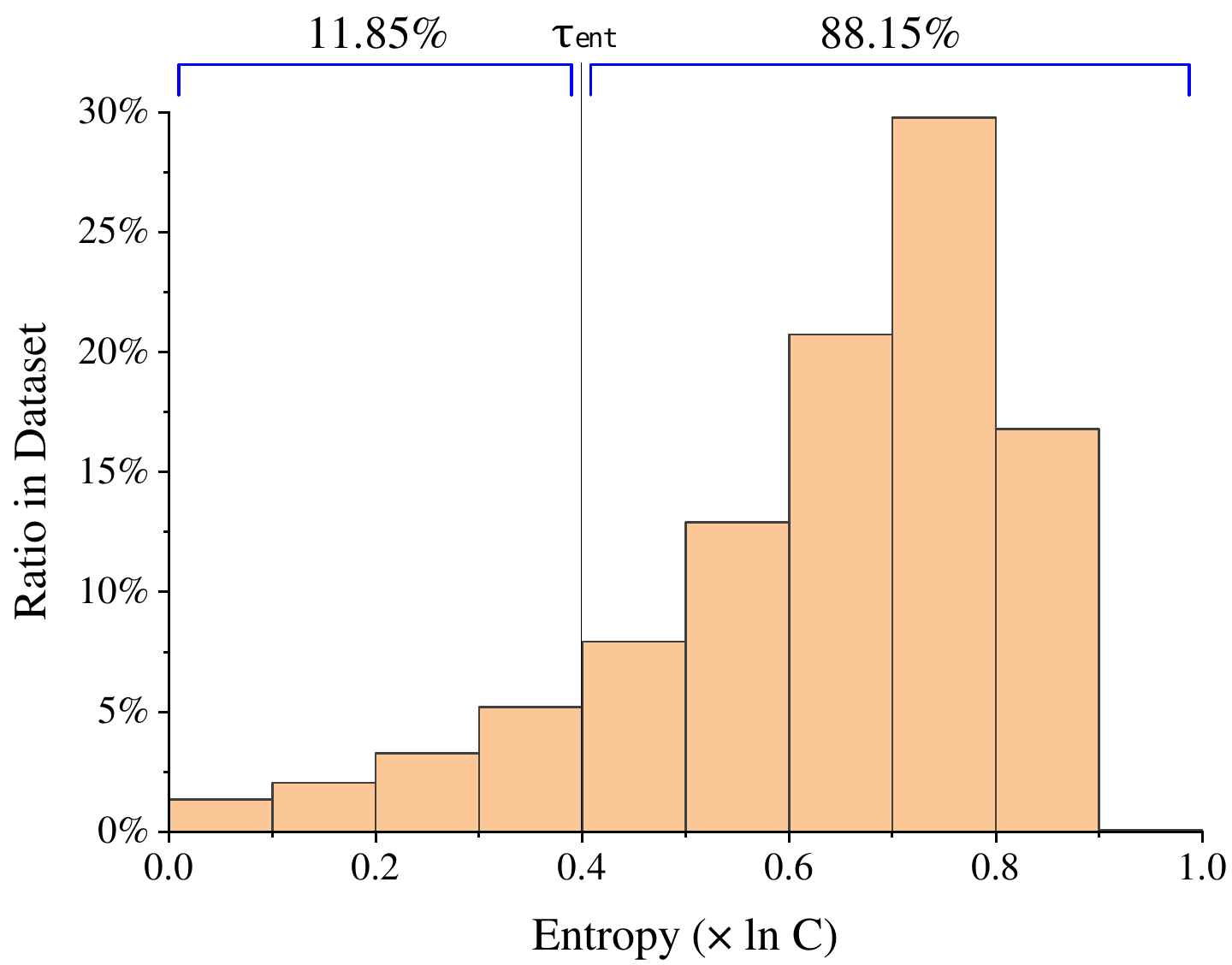}
        \caption{Entropy of samples in ImageNet-C\\\ }
        \label{fig:analysis/entropy}
    \end{subfigure}
    
    \caption{Problem analysis. We use ImageNet pretrained ResNet-50~\cite{he2016deep} and Gaussian noise of level 5 from ImageNet-C~\cite{hendrycks2018benchmarking}. We follow EATA~\cite{niu2022efficient} and SAR~\cite{niu2022towards_SAR} to set the entropy threshold.
    % set the entropy threshold $\tau_{ent}$ as $0.4 \times \log C $, where C is the number of class, following EATA and SAR~\cite{niu2022towards_SAR}. 
    (a) 64.0\% of the classes are selected 5 times or fewer, where each class contains 50 images.
    (b) The less frequently a class is selected, the lower the performance. (c) Only 11.85\% of samples in ImageNet-C are used when entropy based filtering is used.}

    \label{fig:intro}
\end{figure*}

% However, these sample selection based methods are unable to fully utilize the target samples. 
However, the limited amount of sampled data limits the performance improvement.
For instance, only 11.85\% of data from ImageNet-C~\cite{hendrycks2018benchmarking} is selected and utilized by DeYO  to perform naive entropy minimization~\cite{lee2024entropy_deyo}, highlighting the inefficiency in leveraging the available samples.
Furthermore, as depicted in \cref{fig:intro/classes}, 64.0\% of the total classes are sampled less than five times, which leads to poor performance on those classes, as shown in \cref{fig:intro/corr}. 
%Some methods could 
Wang \textit{et al.}~\cite{wang2022continual}
addresses this issue by using consistency
loss,
%~\cite{wang2022continual}, 
which involves comparing predictions with pseudo-labels predicted on augmented images for all samples. While this approach can mitigate the problem, it requires tens of inferences for each sample, rendering it impractical for real-world applications due to its high computational cost.
    
To fully utilize limited samples obtained by entropy-based sample selection, we propose a simple yet effective TTA method, named {F}eature Augmentation based {T}est-time {A}daptation (FATA). 
FATA trains a model by comparing the pseudo-labels of reliable samples, obtained through entropy-based sample selection, with predictions made on the augmented features of these samples using normalization perturbation techniques~\cite{fan2023towards_NP,li2021simple_SFA}. 
Augmented features allow the model to obtain the effect of having more reliable samples with only a small amount of data, thereby acquiring a more generalized representation. 
To augment the features, we adopt Normalization Perturbation (NP)~\cite{fan2023towards_NP}, which randomly perturbs the features in a channel-adaptive manner using channel statistics. 
This enables greater variation in features, not limited to the small variance of the source domain, and facilitates learning from a more diverse set of samples.
Following Stochastic Feature Augmentation~\cite{li2021simple_SFA}, this mechanism is embedded between the last two blocks of the backbone model, mitigating the inefficiency of multiple inferences since only two layers of the model are involved in the process of prediction. FATA can be seamlessly integrated into any method using entropy-based sample selection techniques and is applicable to any architecture.

To validate the effectiveness of FATA, we evaluate our method not only in normal scenarios but also in several challenging scenarios that are likely to occur in the real world, such as label shifts and batch size 1 scenarios, following the settings in SAR~\cite{niu2022towards_SAR}. 
% We utilize ImageNet-C~\cite{hendrycks2018benchmarking} not only in normal scenarios but also in several challenging scenarios, such as label shifts and batch size 1 scenarios.
In our evaluation, the methods incorporating FATA demonstrate superiority in performance, showcasing applicability to various methods and network architectures. Additionally, our meticulously designed ablation studies and analysis illustrate the effectiveness of our components.

\par

Our contributions are summarized as follows:
\begin{itemize}
    \item We analyze and address the problem of data scarcity in sample selection based TTA methods.
    We observe that the majority of data is filtered out and that the number of samples used has a positive correlation with the performance.
    
    \item We propose Feature Augmentation based Test-time Adaptation (FATA), a TTA method that fully exploits limited amount of data. FATA leverages feature augmentation and augmentation loss, and    
    can be seamlessly plugged into any model or TTA method.

    \item We validate FATA on several models and scenarios and demonstrate its effectiveness. FATA outperforms existing methods when plugged into the methods.
\end{itemize}

\section{Related Work}
\label{ch:relatedwork}

\subsection{Test-Time Adaptation}
In order to enable model adaptation in source-free, unlabeled, and online settings, various test-time domain adaptation methodologies~\cite{zhang2022memo,chen2022contrastivetta,niu2022towards_SAR,wang2021tent,niu2022towards_SAR,lim2022ttn,lee2023towards_entropy,park2023label_entropy,wang2023feature} have been introduced, designing unsupervised losses. 
TTA methodologies can be broadly categorized into two approaches: one that utilizes entropy minimization and another that generates reliable pseudo-labels through multiple data augmentation to improve the prediction accuracy.
% that ensures coherence between model predictions through data augmentation-based consistency loss. % another that부터 설명이 이해가 잘 안감. augmentation based pseudo-label에 의존하는게 이 방법론의 방점임.

TENT~\cite{wang2021tent} is the first approach to highlight the test-time adaptation of pre-trained models to given target samples by employing entropy minimization loss. Followed by TENT, several methods~\cite{niu2022efficient, niu2022towards_SAR, lee2024entropy_deyo} employ the entropy minimization.
EATA~\cite{niu2022efficient} suggests sample filtering for reliable adaptation. The authors found that test samples with high entropy lead to noisy gradients, resulting in a  severe performance drop. Consequently, the authors filters out samples with high entropy. 
% Additionally, the authors exclude redundant samples to enhance efficiency.
% EATA~\cite{niu2022efficient} identified that test samples with high entropy lead to noisy gradients, and backpropagation using similar samples only computes redundant gradients, which hinder the adaptation process. Therefore, they utilized reliable low-entropy and non-redundant samples with diverse gradients for efficient model adaptation.
SAR~\cite{niu2022towards_SAR} proposes sharpness-aware and reliable entropy minimization to address the problem of real-world scenarios, such as small batch sizes and online imbalanced label shifts. SAR identifies that samples with large gradient norm also hinder the adaptation process, even if their entropy is low. Based on these observations, SAR minimizes both the sharpness of the entropy loss and the entropy itself, using SAM optimizer~\cite{foret2020sharpness_SAM}. 
DeYO~\cite{lee2024entropy_deyo} observed that using entropy alone as a sample selection criterion is insufficient, as it does not  account for the discriminability of the sampled data, such as structure or shape. The authors demonstrate that samples without the discriminability can be harmful for the adaptation process, even if thy have low entropy. 
To address this problem, DeYO further filters out non-discriminative data from the low entropy samples, using their proposed metric. Although those sample selection based entropy minimization methods have shown promising results, they are limited in performance improvement because they do not utilize the majority of target samples.
% the methods are insufficient because they do not use the majority of target samples, leading to limited performance improvement.

On the other hand, CoTTA~\cite{wang2022continual} augments an input data and performs tens of inferences to generate reliable pseudo-labels, which are then used to train the model. While this method utilizes all the data and addresses the inefficiency problem of entropy minimization-based methods, it is impractical in the real-world applications due to its heavy computational burden.
% more reliable, which are generated by the teacher network, thereby preserving the consistency of predictions and minimizing error accumulation. 
% CoTTA는 2.1 맨 뒤로 보내는게 나을듯.
% limitation 추가. entropy 필터링 방법이 dominant하지만, 이를 사용하면 샘플을 너무 적게 뽑는다. 이를 다 사용 안하는 것은 너무 아깝다. 더 많이 잘 사용하 수 있으면 성능도 오를 것이다. 심지어 deyo같은 경우는 모든 샘플의 9.9프로밖에 사용하지 않는다. cotta같이 pseudo label based방법론들은 잠재적인 해결방법이 될 수는 있겠지만 수십번에 달하는 pseudo label을 얻기 위한 inference 때문에 현실에서는 impractical하다.

\subsection{Feature Augmentation}
In the field of domain generalization, data augmentation has been demonstrated to be an effective method for fully leveraging source data.
% to enhance domain generalization. 
Existing data augmentation methods~\cite{shankar2018generalizing, NEURIPS2018_1d94108e, Zhou_Yang_Hospedales_Xiang_2020} rely on image-space operations, which require careful augmentation design and substantial computational resources.
Recently, feature augmentation has been proposed as a solution to address the limited diversity and inefficiency of data augmentation~\cite{li2021simple_SFA, verma2019manifold,  zhou2021domain}. 
By applying transformations in the feature space~\cite{li2021simple_SFA} to simulate various feature distributions during training~\cite{Tseng2020Cross-Domain}, feature augmentation enhances model generalization to new domains more effectively than traditional data augmentation methods.

% In contrast, feature augmentation addresses these issues by applying transformations in the feature space~\cite{li2021simple_SFA}. By simulating various feature distributions during training~\cite{Tseng2020Cross-Domain}, feature augmentation enhances model generalization to new domains more effectively than traditional data augmentation methods.

% Several methods are based on mixing up latent vectors. For example, 
MixStyle~\cite{zhou2021domain} proposes an explicit augmentation approach that perturbs latent features using domain labels through interpolation. Similarly, Li \textit{et al.} (2021)~\cite{li2021simple_SFA} presents Stochastic Feature Augmentation (SFA), which augments the latent features using a linear function with randomly sampled weights and biases from normal distributions. SFA can be implemented as a plug-in module, making it adaptable for integration into various models.
% Additionally, the study shows that feature augmentation at the later stage of the model is more effective than at the earlier stage.

Meanwhile, Fan \textit{et al.} (2023)~\cite{fan2023towards_NP} propose Normalization Perturbation (NP) inpired by  Adaptive Instance Normalization (AdaIN)~\cite{Huang_2017_ICCV}, a style transfer method that utilizes normalization and transformation of feature channel statistics. Instead of directly perturbing features, NP perturbs the channel statistics to effectively maintain feature content. While feature augmentation techniques have proven effective for domain generalization, their impact on TTA has not been thoroughly explored. Our method integrates feature augmentation into TTA to addresses the data scarcity issues by sampling data based on entropy values.

\section{Problem Analysis}
\label{ch:background}
% \noindent
% This chapter outlines the necessity for further exploitation of the limited sampled data:  \cref{subsec:pre_tta} formulates the TTA problem and introduces how the entropy minimization is formulated in TTA, and \cref{subsec:pre_sampling} describes the existing sampling strategies and the scarcity of sampled data.
% how scarce the data they select is. 

\subsection{Preliminaries}
% \subsubsection{Test-Time Adaptation}
\noindent\textbf{Test-Time Adaptation.}
\label{subsec:pre_tta}
% \input{tab/tta}
% Tab.~\ref{tab:settings} shows the settings of TTA and related adaptation methods. 
Unsupervised domain adaptation (UDA) adapts a model to a target domain without target labels. Source-free domain adaptation removes the necessity of source data from UDA. Test-time training (TTT) adds the constraint of online availability of target data.
% raises the problem of acquiring target data. 
TTT addresses this problem by also training the model online, i.e., in the test time, considering that the target data could be acquired from the real world, in real time. Finally, fully test-time adaptation addresses the most realistic scenario where no source data is available. TTA only adapts a model with unlabeled target data in the test-time.

% \begin{figure}[t!]
%     \centering
%     \includegraphics[width=0.8\linewidth]{fig/an_entropy.pdf}
%     \caption{Entropy of samples in ImageNet-C.}
%     \label{fig:analysis/entropy}
% \end{figure}

% \subsubsection{Entropy Minimization}
\vspace{0.25em}
\noindent\textbf{Entropy Minimization.}
\label{subsec:pre_ent_min}
In TTA scenario, we have a model $f_{\theta}$ with parameter $\theta$ pretrained on the source dataset $\mathcal{D}^\text{train} = \{(\mathbf{x}^\text{train}_i, \mathbf{y}^\text{train}_i)\}_{i=1}^{N^\text{train}}$. Due to the absence of $\mathcal{D}^\text{train}$ and labels $\mathbf{y}^\text{test}$ of the target dataset $\mathcal{D}^\text{test} = \{(\mathbf{x}^\text{test}_i)\}_{i=1}^{N^\text{test}}$ in test time, existing TTA methods have employed unsupervised learning signal. The most dominantly adopted learning signal is Shannon entropy~\cite{shannon2001mathematical}, where the model predicts to minimize the entropy of prediction. Given a model $f_{\theta}$ and a prediction $\hat{y}_i = f_{\theta}(c|\textbf{x})$ according to a class $c$, entropy minimization is formulated as follows:
\begin{equation}
    \min{\text{Ent}_\theta (\textbf{x})}, 
    \text{ where } \text{Ent}_\theta (\mathbf{x}) = - \overset{C}{\underset{i=1}{\sum}} \hat{y}_i \log{\hat{y}_i}, 
\end{equation}
where $C$ is the number of classes.
% $H(\hat{y}) = -\sum_{c}p(\hat{y_c})\text{log}p(\hat{y_c})$ 
% for the probability $\hat{y}_c$ of class $c$. 
% TENT~\cite{wang2021tent} showed that entropy has a negative correlation with an accuracy, making the model more confident. 

\begin{figure*}[ht!]
    \centering
    \includegraphics[width=\linewidth]{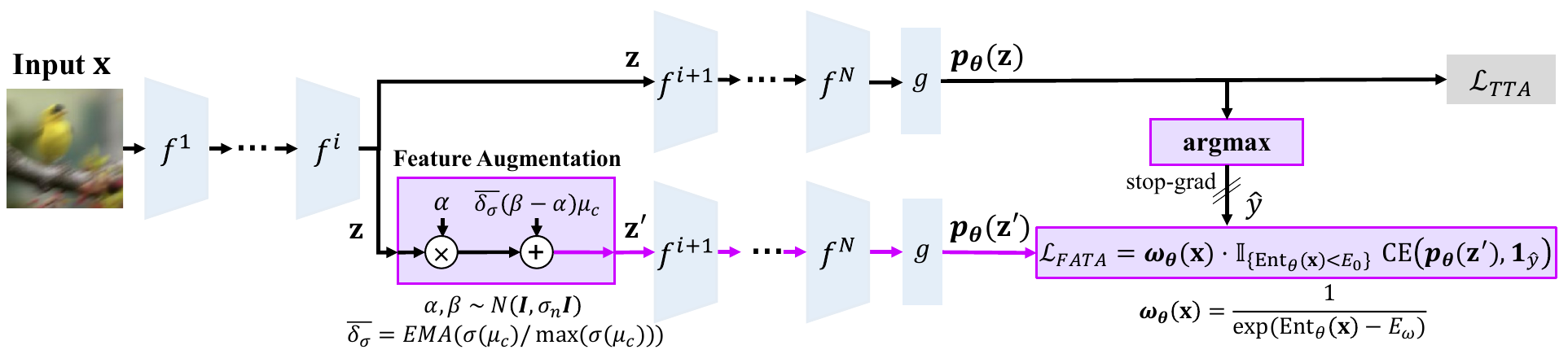}
    \caption{Overview of FATA. There are two prediction branches where one is for obtaining pseudo-label on the reliable data and another is for prediction and updating the model on the augmented feature. We insert the feature augmentation after the $i$-th layer.}
    \label{fig:fta}
\end{figure*}

\vspace{0.25em}
% \subsubsection{Sample Selection Strategy}
\noindent\textbf{Sample Selection Strategy.}
Based on entropy minimization, Niu \textit{et al.}~\cite{niu2022efficient} proposes filtering out unreliable samples by using only samples with low entropy values. 
Therefore, the model minimizes entropy using the samples selected based on sample selection criteria $S(\mathbf{x})$:

\begin{equation}
    \underset{{\theta}}{\text{min}}\ S(\mathbf{x}) \text{Ent}_\theta (\mathbf{x})
    ,\ \ \ \text{where}\ \ S(\mathbf{x}) \overset{\Delta}{=} \mathbb{I}_{\{\mathbf{x} \in \mathcal{S}\}},
\end{equation}
where $\mathbb{I}_{\{\cdot\}}(\cdot)$ is an indicator function and $\mathcal{S}$ is a set of selected samples. For instance, the set for entropy-based sample selection is $\mathcal{S}_\text{ent} = \{\mathbf{x} | \text{Ent}_\theta(\mathbf{x})<\tau_{ent}\}$, where $\tau_{ent}$ is a pre-defined entropy threshold. 

% \begin{equation}
%     \underset{{\theta}}{\text{min}}\ S(\mathbf{x}) \text{Ent}_\theta (\mathbf{x})
%     ,\ \ \ \text{where}\ \ S(\mathbf{x}) \overset{\Delta}{=} \mathbb{I}_{\{\text{Ent}_\theta(\mathbf{x})<\tau_{ent}\}}(\mathbf{x}),
% \end{equation}
% where $\mathbb{I}_{\{\cdot\}}(\cdot)$ is an indicator function and $\tau_{ent}$ is a pre-defined entropy threshold. 

\subsection{Analysis on Sample Selection Strategy}
\label{subsec:pre_sampling}
We measure how many samples are selected by this sampling strategy.
% that EATA, SAR, and DeYO leverage.
We use ResNet50-BN (Batch Normalization) pretrained on ImageNet~\cite{deng2009imagenet}, and count the selected numbers from target dataset $\mathcal{D}^\text{test}$, ImageNet-C~\cite{hendrycks2018benchmarking}, with the entropy threshold $\tau_{ent}$ that the authors of EATA and SAR~\cite{niu2022towards_SAR} recommended.

% \subsubsection{Only a limited samples are used.}
% \noindent\textbf{Only a limited samples are used.}
\vspace{0.25em}
\noindent\textbf{Use of a Limited Number of Samples.}
Limited samples could degrade the performance of adaptation in the real world when the online target data is small or there is a label imbalance.
As shown in \cref{fig:analysis/entropy}, only 11.85\% of $\mathcal{D}^\text{test}$ can be used when the entropy-based filtering strategy is used. This is very small number regarding the number of classes in the dataset, a thousand. This indicates that only five or six samples can be used for adaptation for each class on average (see \cref{fig:intro/classes}), where the accuracies of the less sampled classes are low (see \cref{fig:intro/corr}), leading to poor performance.
% as \cref{fig:intro/classes} shows.
Additionally, DeYO~\cite{lee2024entropy_deyo} selects samples that have helpful structure and shape among the selected sample by entropy-based filtering, resulting in usage of much less samples.

% When the entropy-based filtering strategy is used, only 8.32\% of $\mathcal{D}^\text{test}$ can be used.
%Additionally, DeYO~\cite{lee2024entropy_deyo} selects samples that have helpful structure and shape among the selected sample by entropy-based filtering, resulting in usage of much less samples.

Despite the limited amount of sampled data, these methods naively rely solely on entropy loss without considering generalized representation. We conjecture that this would restrict the exposure to the target domain, leading to limited performance improvement.
To enhance the model's exposure to the target domain by further exploiting reliable sampled data, a more sophisticated method is necessary.
%to enhance the model's exposure to the target domain by further exploiting the reliable sampled data.

\section{Method}
\label{ch:ourmethod}

\subsection{Feature Augmentation}
We augment the target samples to fully exploit the limited amount of data, as shown in \cref{fig:fta}. Instead of conventional data augmentation, we adopt feature augmentation~\cite{li2021simple_SFA}, which allows diverse augmented features. Given an encoder $f$ composed of $N$ layers $f^1, f^2, \cdots, f^N$ and a sample $\mathbf{x} \in \mathbb{R}^{B \times C \times H \times W}$, feature augmentation augments an intermediate feature from the $i$-th layer, $\mathbf{z} = f^i \circ f^{i-1} \circ \cdots \circ f^1(\mathbf{x})$. For example, Normalization Perturbation Plus (NP+)~\cite{fan2023towards_NP} perturbs the channel statistics of an intermediate feature $\mathbf{z} \in \mathbb{R}^{B \times C_i \times H_i \times W_i}$ using normalization for domain generalization as follows:

\begin{equation}
    \mathbf{z'} = \alpha \mathbf{z} + \delta (\beta - \alpha) \mu_c ,
    \label{eq:NP+}
\end{equation}
where $\alpha$, $\beta \in \mathbb{R}^{B\times C}$ are the random noises sampled from $N(\mathbf{I}, \sigma_{n}\mathbf{I})$, $\delta = \text{Var}(\mu_c) / \text{max}(\text{Var}(\mu_c))$ is the normalized variance, $\mu_c = \{ \mu_c^j \}_{j=1}^{B} \in \mathbb{R}^{1 \times C_i}$ is the channel-wise feature mean. 

FATA augments an intermediate feature $\mathbf{z}$ as follows:
\begin{equation}
    \mathbf{z'} = \alpha \mathbf{z} + \overline{\delta_\sigma} (\beta - \alpha) \mu_c ,
\end{equation}
where
$\overline{\delta_\sigma}$ is the exponential moving average of the normalized standard deviation $\delta_\sigma = \sigma(\mu_c) / \text{max}(\sigma(\mu_c))$ and $\sigma$ is the standard deviation operator. We replace $\delta$ with $\overline{\delta_\sigma}$ to address the following issues. 
Firstly, unlike domain generalization, TTA adapts to a specific domain from a limited amount of data. To adaptively adjust the noise to fit the target domain, we add an exponential moving average that estimates the statistics of the target domain.
Secondly, \cref{eq:NP+} introduces the variance with the magnitude of square of variance, as the normalized statistic variance $\delta$ adjusts the channel mean $\mu_c$ to control the random noise for each channel.
To address this issue, we replace the variance to the standard deviation.

%in Eq.~\ref{eq:NP+}, which is identical to $\alpha \sigma_c \cdot {(x - \delta\mu_c)}/{\sigma_c} + \delta\beta \mu_c $, 
% the normalized statistic variance $\delta$ adjusts the channel mean $\mu_c$ to control the random noise for each channel, introducing the variance with the magnitude of square of variance. To address this issue, I replace the variance to the standard deviation.

\begin{table*}[th!]
\centering
\def\arraystretch{1.25}
\setlength{\tabcolsep}{2pt}
\begin{subtable}[t]{\textwidth}
\centering
{\scriptsize
\begin{tabularx}{\textwidth}{c|YYY|YYYY|YYYY|YYYY|Yc}
\toprule
\centering
\multirow{2}{*}{Method} & 
\multicolumn{3}{c|}{Noise} & \multicolumn{4}{c|}{Blur} & \multicolumn{4}{c|}{Weather} & \multicolumn{4}{c|}{Digital} & \multirow{2}{*}{ Avg.} & \multirow{2}{*}{$\Delta$Perf.}
\\[-0.7ex]
& {Gauss.} &  {Shot} &  {Impul.} &  {Defoc.} &  {Glass} &  {Motion} &  {Zoom} &  {Snow} &  {Frost} &  {Fog} &  {Brit.} &  {Contr.} &  {Elastic} &  {Pixel} &  {JPEG}  && \\
\hline
No Adapt & 16.06 & 16.76 & 16.72 & 14.94 & 15.36 & 26.23 & 38.87 & 34.29 & 33.21 & 47.75 & 65.35 & 16.91 & 43.97 & 49.04 & 40.01 & 31.70 & - \\[-0.3ex]
\hline
TENT~\cite{wang2021tent} & 29.36 & 31.30 & 30.13 & 28.22 & 27.88 & 41.40 & 49.30 & 47.36 & 41.73 & 57.53 & 67.50 & 30.08 & 54.90 & 58.58 & 52.57 & 43.19 & - \\[-0.3ex]
% \rowcolor{lightgray}
% TENT+FATA & 35.03 & 37.23 & 36.07 & 33.16 & 32.93 & 47.06 & 51.46 & 51.30 & 45.10 & 59.33 & 67.26 & 45.14 & 57.30 & 60.16 & 54.87 & 47.56 & +4.37\\ [-0.3ex]

EATA~\cite{niu2022efficient} & 34.58 & 37.34 & 35.87 & 33.55 & 33.24 & 47.40 & 52.90 & 51.72 & 45.71 & 59.86 & 68.12 & 44.65 & 57.90 & 60.39 & 55.03 & 47.88 & - \\[-0.3ex]
\rowcolor{lightgray}
EATA+FATA & 35.31 & 38.00 & 36.31 & 34.63 & 34.32 & 48.15 & 52.44 & 52.17 & 46.26 & 59.93 & 67.43 & 46.48 & 57.77 & 60.19 & 55.05 & 48.29 & +0.41 \\ [-0.3ex]

SAR~\cite{niu2022towards_SAR} & 29.99 & 31.98 & 30.95 & 28.33 & 26.11 & 42.01 & 49.51 & 47.63 & 42.61 & 57.70 & 67.37 & 39.46 & 54.58 & 58.62 & 52.64 & 43.97 & - \\[-0.3ex]
\rowcolor{lightgray}
SAR+FATA & 35.00 & 37.05 & 35.70 & 33.55 & 32.61 & 47.35 & 51.55 & 51.23 & 45.29 & 59.33 & 67.08 & 42.23 & 57.30 & 60.06 & 54.72 & 47.34 & +3.40\\ [-0.3ex]

DeYO~\cite{lee2024entropy_deyo} & 35.68 & 38.19 & 37.39 & 33.99 & 33.65 & 48.27 & 52.94 & 52.34 & 46.32 & 60.50 & 68.01 & 44.34 & 58.25 & 61.16 & 55.58 & 48.44 & - \\ [-0.3ex]
\rowcolor{lightgray}
DeYO+FATA & 37.06 & 38.76 & 38.05 & 34.80 & 34.59 & 49.16 & 52.91 & 52.82 & 46.78 & 60.67 & 67.73 & 47.67 & 58.47 & 61.19 & 55.69 & 49.09 & +0.65\\ [-0.3ex]
\bottomrule
\end{tabularx}
}
\caption{ResNet50 (BN)
\vspace{0.1em}}
\end{subtable}

\begin{subtable}[t]{\textwidth}
\centering
{\scriptsize
\begin{tabularx}{\textwidth}{c|YYY|YYYY|YYYY|YYYY|Yc}
\toprule
\centering
\multirow{2}{*}{Method} & 
\multicolumn{3}{c|}{Noise} & \multicolumn{4}{c|}{Blur} & \multicolumn{4}{c|}{Weather} & \multicolumn{4}{c|}{Digital} & \multirow{2}{*}{ Avg.} & \multirow{2}{*}{$\Delta$Perf.}
\\[-0.7ex]
& {Gauss.} &  {Shot} &  {Impul.} &  {Defoc.} &  {Glass} &  {Motion} &  {Zoom} &  {Snow} &  {Frost} &  {Fog} &  {Brit.} &  {Contr.} &  {Elastic} &  {Pixel} &  {JPEG}  && \\
\hline

No Adapt & 17.98 & 19.83 & 17.87 & 19.75 & 11.35 & 21.41 & 24.91 & 40.43 & 47.31 & 33.60 & 69.28 & 36.27 & 18.61 & 28.40 & 52.28 & 30.62 & - \\ [-0.3ex]
\hline
TENT~\cite{wang2021tent} & 6.45 & 9.12 & 8.11 & 16.62 & 12.66 & 25.54 & 28.93 & 31.87 & 39.63 & 4.50 & 71.11 & 43.81 & 15.81 & 49.57 & 55.59 & 27.95 & - \\[-0.3ex]
% \rowcolor{lightgray}
% TENT+FATA & 39.53 & 42.38 & 40.24 & 24.30 & 24.96 & 37.28 & 40.00 & 52.14 & 51.31 & 1.88 & 73.41 & 50.24 & 10.21 & 56.20 & 57.52 & 40.11 & +12.16\\ [-0.3ex]

EATA~\cite{niu2022efficient} & 37.16 & 40.33 & 38.64 & 28.98 & 27.58 & 36.69 & 38.72 & 51.25 & 49.47 & 55.12 & 71.98 & 49.77 & 41.39 & 55.90 & 57.90 & 45.39 & - \\[-0.3ex]
\rowcolor{lightgray}
EATA+FATA & 42.67 & 45.64 & 43.92 & 32.53 & 31.07 & 42.96 & 46.72 & 57.03 & 54.81 & 61.80 & 73.81 & 54.56 & 51.12 & 60.94 & 60.89 & 50.70 & +5.31\\ [-0.3ex]

SAR~\cite{niu2022towards_SAR} & 28.56 & 30.96 & 29.85 & 18.39 & 18.15 & 30.67 & 30.70 & 41.97 & 43.76 & 6.20 & 70.75 & 44.08 & 15.50 & 48.94 & 55.28 & 34.25 & - \\[-0.3ex]
\rowcolor{lightgray}
SAR+FATA & 40.15 & 42.46 & 40.85 & 24.99 & 25.21 & 38.33 & 40.46 & 52.90 & 50.85 & 0.23 & 73.47 & 49.92 & 40.17 & 55.94 & 57.31 & 42.22 & +7.97\\ [-0.3ex]

DeYO~\cite{lee2024entropy_deyo} & 39.46 & 41.90 & 41.03 & 22.27 & 24.11 & 38.48 & 37.87 & 50.51 & 49.59 & 1.43 & 73.17 & 49.95 & 41.54 & 55.96 & 57.82 & 41.67 & - \\ [-0.3ex]
\rowcolor{lightgray}
DeYO+FATA & 39.71 & 42.49 & 41.37 & 22.29 & 24.46 & 38.90 & 38.31 & 51.23 & 50.02 & 56.31 & 73.19 & 50.03 & 42.10 & 55.99 & 57.79 & 45.61 & +3.97 \\ [-0.3ex]

\bottomrule
\end{tabularx}
}
\caption{ResNet50 (GN)
\vspace{0.1em}
}
\end{subtable}

\begin{subtable}[t]{\textwidth}
\centering
{\scriptsize
\begin{tabularx}{\textwidth}{c|YYY|YYYY|YYYY|YYYY|Yc}
\toprule
\centering
\multirow{2}{*}{Method} & 
\multicolumn{3}{c|}{Noise} & \multicolumn{4}{c|}{Blur} & \multicolumn{4}{c|}{Weather} & \multicolumn{4}{c|}{Digital} & \multirow{2}{*}{ Avg.} & \multirow{2}{*}{$\Delta$Perf.}
\\[-0.7ex]
& {Gauss.} &  {Shot} &  {Impul.} &  {Defoc.} &  {Glass} &  {Motion} &  {Zoom} &  {Snow} &  {Frost} &  {Fog} &  {Brit.} &  {Contr.} &  {Elastic} &  {Pixel} &  {JPEG}  && \\
\hline
No Adapt & 9.48 & 6.75 & 8.23 & 28.99 & 23.45 & 33.87 & 27.13 & 15.91 & 26.49 & 47.18 & 54.66 & 44.09 & 30.55 & 44.50 & 47.80 & 29.94 & - \\ [-0.3ex]
\hline
TENT~\cite{wang2021tent} & 42.59 & 1.92 & 43.80 & 52.49 & 48.43 & 55.72 & 51.01 & 30.03 & 24.92 & 66.67 & 74.90 & 64.66 & 53.72 & 67.00 & 64.27 & 49.48 & - \\[-0.3ex]
% \rowcolor{lightgray}
% TENT+FATA & 45.34 & 44.36 & 46.40 & 53.64 & 51.12 & 57.28 & 53.54 & 60.73 & 60.09 & 68.60 & 75.55 & 64.91 & 60.10 & 68.82 & 65.14 & 58.37 & +8.89\\ [-0.3ex]

EATA~\cite{niu2022efficient} & 50.23 & 49.70 & 51.33 & 55.32 & 55.44 & 59.64 & 57.00 & 63.29 & 62.38 & 70.85 & 75.95 & 66.95 & 64.16 & 69.39 & 67.51 & 61.28 & - \\[-0.3ex]
\rowcolor{lightgray}
EATA+FATA & 53.15 & 53.21 & 54.05 & 57.23 & 57.84 & 62.08 & 60.78 & 66.76 & 65.41 & 72.10 & 76.89 & 67.65 & 67.13 & 71.68 & 69.01 & 63.67 & +2.39\\ [-0.3ex]

SAR~\cite{niu2022towards_SAR} & 44.03 & 42.96 & 45.54 & 53.07 & 49.81 & 55.78 & 51.44 & 58.00 & 55.43 & 66.34 & 74.58 & 64.14 & 55.02 & 66.84 & 64.06 & 56.47 & - \\[-0.3ex]
\rowcolor{lightgray}
SAR+FATA & 51.67 & 3.69 & 52.48 & 57.00 & 56.93 & 61.42 & 59.76 & 65.70 & 64.70 & 71.33 & 76.21 & 64.42 & 66.43 & 71.70 & 68.30 & 59.45 & +2.98\\ [-0.3ex]

DeYO~\cite{lee2024entropy_deyo} & 48.56 & 47.66 & 53.66 & 58.32 & 58.53 & 63.05 & 59.96 & 67.12 & 65.85 & 73.23 & 77.97 & 67.98 & 67.87 & 73.17 & 69.90 & 63.52  & - \\ [-0.3ex]
\rowcolor{lightgray}
DeYO+FATA & 46.83 & 53.78 & 54.2 & 58.56 & 58.57 & 63.37 & 60.16 & 67.46 & 66.24 & 73.46 & 78.19 & 68.55 & 67.78 & 73.33 & 69.93 & 64.03 & +0.51\\ [-0.3ex]

\bottomrule
\end{tabularx}
}
\caption{ViT-B (LN)
}
\end{subtable}

\caption{Image classification results on ImageNet-C~\cite{hendrycks2018benchmarking}. ResNet50 with BN/GN and ViT-B with LN are used for this experiment. We use the accuracy (\%) as the metric. 
$\Delta$Perf. is the performance gap between methods without and with FATA.
% The {best} and \underline{second-best} results are highlighted.
}
\label{tab:imagenet-c}
% Best results are {highlighted} and second-best results are \underline{underlined}.}
\end{table*}

\subsection{FATA Loss}
In order to fully utilize the augmented features, we propose the FATA loss, an augmentation loss that is applied to augmented features.

\vspace{0.25em}
% \subsubsection{Augmentation Loss}
\noindent\textbf{Augmentation Loss.}
Given a classifier $g$, the output probability 
$\mathbf{p}_{\mathbf{\theta}}(\mathbf{z}) = g \circ f^N \circ f^{N-1} \circ \cdots \circ f^{i+1}(\mathbf{z})$. We propose an augmentation loss based on cross-entropy between an output of the augmented feature and a pseudo-label from the original feature. Given the pseudo-label $\hat{y} = \text{stopgrad}({\text{argmax}}(\mathbf{p_{\mathbf{\theta}}}(\mathbf{z})))$, our augmentation loss is formulated as follows:

\begin{equation}
    \mathcal{L}_\text{aug}(\mathbf{x}; \mathbf{\theta}) = \text{CE}(\mathbf{p}_\mathbf{\theta}(\mathbf{z'}), \mathbf{1}_{\hat{y}}).
\end{equation}
Unlike CoTTA~\cite{wang2022continual}, FATA updates a model on the augmented features and uses a prediction on the original data as a pseudo-label.
Therefore, the model can make predictions on more diverse features and be updated on those features.
Also, the pseudo-label is reliable because the data has already been sampled by the entropy threshold $E_0$.

\vspace{0.25em}
\noindent\textbf{Sample Selection and Weighting.}
Following EATA~\cite{niu2022efficient}, we apply entropy-based sample selection and sample weighting. Given an entropy threshold $E_0$ and a normalization factor $E_w$, the sample selection criteria is $\{ \mathbf{x} | \text{Ent}_\mathbf{\theta}(\mathbf{x}) < E_0 \}$ and the sample weighting function $\mathbf{\omega}_{\mathbf\theta}$ is formulated as 
$
    \mathbf{\omega}_{\mathbf{\theta}}(\mathbf{x}) = {1}/{\exp{(\text{Ent}_\mathbf{\theta}(\mathbf{x}) - E_w)}}
$.

\vspace{0.25em}
\noindent\textbf{FATA Augmentation Loss.}
Finally, we incorporate sample selection and weighting to our augmentation loss as follows:
\begin{equation}
    \mathcal{L}_\text{FATA}(\mathbf{x}; \mathbf{\theta}) = \mathbf\omega_\mathbf{\theta}(\mathbf{x}) \cdot \mathbb{I}_{\{\text{Ent}_\mathbf{\theta}(\mathbf{x}) < E_0\}} \text{CE}(\mathbf{p}_\mathbf{\theta}(\mathbf{z'}), \mathbf{1}_{\hat{y}}),
\end{equation}
where $\text{CE}(p, q)$ is the cross-entropy function.

\vspace{0.25em}
% \subsubsection{Total Loss}
\noindent\textbf{Total Loss.}
Given a TTA loss $\mathcal{L}_\text{TTA}$ such as entropy minimization loss, we incorporate the FATA augmentation loss to the TTA loss. Consequently, the total loss $\mathcal{L}$ is as follows:
\begin{equation}
    \mathcal{L} = \mathcal{L}_\text{TTA} + \mathcal{L}_\text{FATA},
\end{equation}
which combines TTA loss and FATA augmentation loss. The proposed loss can be plugged into any method, without modifying the TTA loss such as sample selection based entropy minimization loss.
\section{Experiment}
\label{sec:exp}
\noindent
\subsection{Experimental Settings}

\begin{table*}[t!]
\centering
% \newcolumntype{Y}{>{\centering\arraybackslash}X}
\def\arraystretch{1.25}
\setlength{\tabcolsep}{2pt}
\scriptsize{

\begin{subtable}[t]{\textwidth}
\centering
\begin{tabularx}{\textwidth}{c|*{3}{Y}|*{4}{Y}|*{4}{Y}|*{4}{Y}|Yc}
% \hline
\toprule
% \hline
% Time & \multicolumn{15}{l|}{$\; t \xrightarrow{\hspace{1.55\columnwidth}}$} & \multirow{2}{*}{Mean}  \\
% \cline{1-16}
\multirow{2}{*}{Method} & 
\multicolumn{3}{c|}{Noise} & \multicolumn{4}{c|}{Blur} & \multicolumn{4}{c|}{Weather} & \multicolumn{4}{c|}{Digital} & \multirow{2}{*}{ Avg.} & \multirow{2}{*}{$\Delta$Perf.}
\\[-0.7ex]
& {Gauss.} &  {Shot} &  {Impul.} &  {Defoc.} &  {Glass} &  {Motion} &  {Zoom} &  {Snow} &  {Frost} &  {Fog} &  {Brit.} &  {Contr.} &  {Elastic} &  {Pixel} &  {JPEG}  &  &  \\
\hline
No Adapt & 17.98 & 19.84 & 17.88 & 19.75 & 11.35 & 21.42 & 24.92 & 40.43 & 47.30 & 33.59 & 69.28 & 36.27 & 18.61 & 28.40 & 52.28 & 30.62 & - \\[-0.3ex]
\hline
TENT~\cite{wang2021tent} & 3.36 & 4.31 & 4.18 & 16.75 & 3.49 & 28.00 & 29.36 & 18.65 & 24.72 & 2.21 & 72.03 & 46.17 & 8.12 & 52.43 & 56.25 & 24.67 & - \\[-0.3ex]
% \rowcolor{lightgray}
% TENT+FATA & 35.03 & 37.23 & 36.07 & 33.16 & 32.93 & 47.06 & 51.46 & 51.30 & 45.10 & 59.33 & 67.26 & 45.14 & 57.30 & 60.16 & 54.87 & 47.56 & +4.37\\ [-0.3ex]

EATA~\cite{niu2022efficient} & 24.68 & 28.01 & 25.53 & 17.75 & 17.43 & 28.68 & 29.20 & 44.52 & 44.34 & 41.92 & 70.93 & 44.86 & 27.53 & 45.90 & 55.62 & 36.46 & - \\[-0.3ex]
\rowcolor{lightgray}
EATA+FATA & 26.17 & 31.42 & 27.12 & 20.34 & 17.09 & 32.20 & 23.94 & 49.54 & 50.02 & 11.05 & 72.86 & 49.64 & 7.40 & 51.85 & 58.38 & 35.27 & -1.19 \\ [-0.3ex]

SAR~\cite{niu2022towards_SAR} & 23.35 & 26.27 & 23.82 & 18.71 & 15.54 & 28.78 & 30.62 & 45.55 & 44.93 & 25.58 & 72.18 & 44.56 & 15.04 & 47.22 & 56.05 & 34.55 & - \\[-0.3ex]
\rowcolor{lightgray}
SAR+FATA & 34.91 & 39.20 & 36.45 & 24.14 & 22.33 & 36.89 & 39.47 & 54.26 & 51.55 & 8.06 & 73.88 & 50.98 & 41.29 & 55.74 & 58.54 & 41.85 & +7.30\\ [-0.3ex]

DeYO~\cite{lee2024entropy_deyo} & 41.34 & 44.11 & 42.69 & 22.39 & 24.22 & 41.43 & 28.93 & 54.06 & 51.79 & 2.14 & 73.17 & 53.42 & 47.84 & 59.86 & 59.67 & 43.14 & - \\ [-0.3ex]
\rowcolor{lightgray}
DeYO+FATA & 42.06 & 44.52 & 42.52 & 26.75 & 27.33 & 42.48 & 43.31 & 56.42 & 54.07 & 2.58 & 73.97 & 54.10 & 48.28 & 60.21 & 60.39 & 45.27 & +2.13 \\ [-0.3ex]

% \rowcolor{lightgray}
% Ours & - \\ [-0.3ex]
\bottomrule
\end{tabularx}
\caption{ResNet50 (GN)
\vspace{0.2em}}

\end{subtable}

% \scriptsize{
\begin{subtable}[t]{\textwidth}
\centering
\begin{tabularx}{\textwidth}{c|*{3}{Y}|*{4}{Y}|*{4}{Y}|*{4}{Y}|Yc}
% \hline
\toprule
% \hline
% Time & \multicolumn{15}{l|}{$\; t \xrightarrow{\hspace{1.55\columnwidth}}$} & \multirow{2}{*}{Mean}  \\
% \cline{1-16}
\multirow{2}{*}{Method} & 
\multicolumn{3}{c|}{Noise} & \multicolumn{4}{c|}{Blur} & \multicolumn{4}{c|}{Weather} & \multicolumn{4}{c|}{Digital} & \multirow{2}{*}{ Avg.} & \multirow{2}{*}{$\Delta$Perf.}
\\[-0.7ex]
& {Gauss.} &  {Shot} &  {Impul.} &  {Defoc.} &  {Glass} &  {Motion} &  {Zoom} &  {Snow} &  {Frost} &  {Fog} &  {Brit.} &  {Contr.} &  {Elastic} &  {Pixel} &  {JPEG}  &  &  \\
\hline

No Adapt & 9.48 & 6.75 & 8.23 & 28.99 & 23.45 & 33.87 & 27.13 & 15.91 & 26.49 & 47.18 & 54.66 & 44.09 & 30.55 & 44.50 & 47.80 & 29.94 & - \\ [-0.3ex]
\hline
TENT~\cite{wang2021tent} & 42.72 & 1.40 & 44.53 & 52.49 & 48.67 & 56.10 & 51.09 & 22.96 & 20.87 & 66.78 & 75.01 & 64.98 & 53.96 & 67.03 & 64.38 & 48.86 & - \\[-0.3ex]
% \rowcolor{lightgray}
% TENT+FATA & 39.53 & 42.38 & 40.24 & 24.30 & 24.96 & 37.28 & 40.00 & 52.14 & 51.31 & 1.88 & 73.41 & 50.24 & 10.21 & 56.20 & 57.52 & 40.11 & +12.16\\ [-0.3ex]

EATA~\cite{niu2022efficient} & 33.54 & 25.16 & 31.98 & 44.70 & 38.86 & 47.11 & 42.37 & 39.67 & 40.81 & 61.14 & 67.81 & 61.79 & 47.51 & 59.00 & 59.18 & 46.71 & - \\[-0.3ex]
\rowcolor{lightgray}
EATA+FATA & 39.33 & 1.29 & 37.47 & 48.06 & 43.99 & 51.20 & 47.29 & 48.75 & 47.02 & 66.32 & 71.84 & 64.09 & 56.50 & 63.51 & 62.94 & 49.97 & +3.12\\ [-0.3ex]

SAR~\cite{niu2022towards_SAR} & 41.29 & 36.39 & 42.08 & 53.54 & 50.68 & 57.62 & 52.78 & 58.68 & 44.52 & 68.61 & 75.90 & 65.46 & 57.86 & 68.79 & 65.91 & 56.01 & - \\[-0.3ex]
\rowcolor{lightgray}
SAR+FATA & 50.28 & 48.57 & 51.20 & 57.08 & 56.69 & 61.94 & 59.65 & 65.79 & 64.58 & 71.64 & 76.67 & 66.78 & 66.40 & 72.06 & 68.13 & 62.50 & +5.46\\ [-0.3ex]

DeYO~\cite{lee2024entropy_deyo} & 53.29 & 53.92 & 54.22 & 58.85 & 59.34 & 64.45 & 49.76 & 68.06 & 66.63 & 73.71 & 78.23 & 68.15 & 68.75 & 73.76 & 70.74 & 64.12 & - \\ [-0.3ex]
\rowcolor{lightgray}
DeYO+FATA & 52.87 & 52.89 & 53.87 & 58.34 & 58.60 & 63.84 & 61.03 & 67.56 & 65.91 & 72.66 & 77.49 & 67.64 & 68.15 & 72.96 & 69.81 & 64.24 & +0.12 \\ [-0.3ex]

% \rowcolor{lightgray}
% Ours & {55.2} & {50.5} & {65.6} & {38.1} & {46.1} & {29.3} & {32.6} & {23.4} & {33.1} & {21.8} & {9.1} & {38.5} & {24.5} & {44.3} & {28.9} & {36.1} \\ [-0.3ex]

\bottomrule
\end{tabularx}
\caption{ViT-B (LN)}
\end{subtable}
}
\caption{Image classification results on ImageNet-C~\cite{hendrycks2018benchmarking} under batch size 1. ResNet50 with GN and ViT-B with LN are used for this experiment. We use the accuracy (\%) as the metric. 
$\Delta$Perf. is the performance gap between the original method and another version where our method is incorporated.
% The {best} and \underline{second-best} results are highlighted.
}
\label{tab:bs1}
% 
% Best results are {highlighted} and second-best results are \underline{underlined}.}
\end{table*}

\noindent\textbf{Benchmark and Test Scenarios.}
We use the ImageNet-C~\cite{hendrycks2018benchmarking} and the Office-Home~\cite{venkateswara2017deep_OfficeHome} datasets for our evaluation. ImageNet-C includes corrupted images from the ImageNet dataset across 15 different corruption types. The dataset contains 50,000 images for each corruption type, resulting in a total of 750,000 images. Office-Home includes 15,588 images of 65 classes from 4 domains. Following the scenarios outlined by Niu \textit{et al.}~\cite{niu2022towards_SAR}, we validate the effectiveness and robustness of our method in three scenarios: 1) In the normal scenario, a model adapts to streaming corrupted input where the label distribution is balanced, allowing the use of a large batch size; 2) In the batch size of one scenario, a model is exposed to a single image per iteration; 3) In the online imbalanced label distribution shift scenario, the labels within a batch are highly imbalanced. 

% normal, wild test scenario (bs1, label shift)

\vspace{0.25em}
% \noindent\textbf{Models.}
\noindent\textbf{Models.}
To demonstrate the applicability of our method to various models, we use models incorporating three different normalization layers in our experiments: Batch Normalization (BN), Group Normalization (GN), and Layer Normalization (LN). For BN and GN, we use ResNet-50, and for LN, we use the ViT-Base model. For ImageNet-C, all the models are pretrained on ImageNet dataset and adapted at test-time. For Office-Home, the models are pretrained on a source domain and adapted to a target domain at test-time.

\vspace{0.25em}
\noindent\textbf{Baselines.}
We compare our method to state-of-the-art fully test-time adaptation methods, TENT~\cite{wang2021tent}, 
EATA~\cite{niu2022efficient},  
SAR~\cite{niu2022towards_SAR}, and
DeYO~\cite{lee2024entropy_deyo}. 
Since our method is proposed to address data scarcity when sample filtering is used, we integrate our method with EATA, SAR, and DeYO, which all employ filtering strategies.
We further compare our method to MEMO~\cite{zhang2022memo} and CoTTA~\cite{wang2022continual} on Office-Home.

\vspace{0.25em}
\noindent\textbf{Implementation Details.}
We implement FATA on the PyTorch framework. We set $E_0$ to $0.5\ln{C}$ and $E_\omega$ to $0.4\ln{C}$, following the setting of DeYO~\cite{lee2024entropy_deyo}, where $C$ is the number of classes in the target dataset. In addition, we use the same optimizer as the existing methods with which we integrate FATA. 
We set the layer to inject the feature augmentation $i$ to 3 for ResNet50 and 11 for ViT-B. We set the standard deviation for the noise $\sigma_n$ to 1.0 and the smoothing factor for exponential moving average $\lambda_{EMA}$ to 0.95.
We use the default batch size of 64. We set the learning rate as 0.0005 and 0.001 for ResNet50 and ViT-B, respectively. When the batch size is one, we set the learning rate as 0.00025 and 0.000016 for ResNet50 and ViT-B, respectively.

\subsection{Results}
% \subsubsection{Comparison on Normal Scenario}
\noindent\textbf{Comparison on Normal Scenario.}
\cref{tab:imagenet-c} shows the comparison of performance in the normal scenario. The results clearly indicate that integrating our proposed method, FATA, yields superior results across various architectures and methods, compared to the original versions. Notably, the enhancement is most significant when ResNet-50 with GN is utilized, showcasing an impressive average improvement of up to 7.97 points. This substantial gain highlights the effectiveness of FATA, demonstrating its capability regardless of the underlying model architecture or TTA method. Furthermore, the consistent improvements across different settings demonstrate the robustness of FATA.

\begin{table*}[t]
\centering
\def\arraystretch{1.25}
\setlength{\tabcolsep}{2pt}
\scriptsize {
\begin{tabularx}{\textwidth}{c|*{3}{Y}|*{4}{Y}|*{4}{Y}|*{4}{Y}|Yc}
% \hline
\toprule
% \hline
% Time & \multicolumn{15}{l|}{$\; t \xrightarrow{\hspace{1.55\columnwidth}}$} & \multirow{2}{*}{Mean}  \\
% \cline{1-16} 
\multirow{2}{*}{Method} & 
\multicolumn{3}{c|}{Noise} & \multicolumn{4}{c|}{Blur} & \multicolumn{4}{c|}{Weather} & \multicolumn{4}{c|}{Digital} & \multirow{2}{*}{ Avg.} & \multirow{2}{*}{$\Delta$Perf.}
\\[-0.7ex]
& {Gauss.} &  {Shot} &  {Impul.} &  {Defoc.} &  {Glass} &  {Motion} &  {Zoom} &  {Snow} &  {Frost} &  {Fog} &  {Brit.} &  {Contr.} &  {Elastic} &  {Pixel} &  {JPEG}  &  &  \\
\hline
% \midrule
No Adapt & 17.87 & 19.91 & 17.90 & 19.70 & 11.21 & 21.28 & 24.86 & 40.38 & 47.40 & 33.57 & 69.24 & 36.27 & 18.65 & 28.34 & 52.17 & 30.58 & - \\[-0.3ex]
\hline
TENT~\cite{wang2021tent} & 3.54 & 4.17 & 3.78 & 16.73 & 7.53 & 25.73 & 31.50 & 20.14 & 29.51 & 2.39 & 72.17 & 46.09 & 8.38 & 52.11 & 56.22 & 25.33 & - \\[-0.3ex]
% \rowcolor{lightgray}
% TENT+FATA & 35.03 & 37.23 & 36.07 & 33.16 & 32.93 & 47.06 & 51.46 & 51.30 & 45.10 & 59.33 & 67.26 & 45.14 & 57.30 & 60.16 & 54.87 & 47.56 & +4.37\\ [-0.3ex]

EATA~\cite{niu2022efficient} & 25.26 & 29.49 & 27.72 & 14.90 & 16.61 & 24.41 & 28.60 & 34.89 & 29.71 & 41.41 & 62.65 & 35.22 & 26.14 & 41.70 & 48.69 & 32.49 & - \\[-0.3ex]
\rowcolor{lightgray}
EATA+FATA & 41.75 & 43.98 & 41.78 & 27.95 & 27.63 & 42.47 & 45.80 & 56.82 & 53.83 & 62.12 & 73.32 & 54.33 & 50.38 & 61.42 & 60.52 & 49.61 & +17.12 \\ [-0.3ex]

SAR~\cite{niu2022towards_SAR} & 33.81 & 36.90 & 33.90 & 18.35 & 20.48 & 33.13 & 33.97 & 29.87 & 45.08 & 2.77 & 71.95 & 46.78 & 7.43 & 51.91 & 56.06 & 34.83 & - \\[-0.3ex]
\rowcolor{lightgray}
SAR+FATA & 42.97 & 45.32 & 43.91 & 28.29 & 27.76 & 41.51 & 44.26 & 55.21 & 53.72 & 1.56 & 73.78 & 53.27 & 2.43 & 60.14 & 59.47 & 42.24 & +7.41\\ [-0.3ex]

DeYO~\cite{lee2024entropy_deyo} & 40.84 & 18.13 & 3.46 & 22.63 & 23.64 & 41.10 & 7.11 & 52.57 & 51.39 & 58.72 & 73.12 & 52.21 & 46.38 & 59.09 & 59.43 & 40.66 & - \\ [-0.3ex]
\rowcolor{lightgray}
DeYO+FATA & 44.30 & 46.29 & 1.06 & 25.89 & 29.58 & 45.00 & 5.06 & 58.27 & 54.69 & 62.98 & 73.85 & 55.52 & 53.23 & 62.52 & 61.19 & 45.30 & +4.64 \\ [-0.3ex]

\bottomrule
\end{tabularx}
}
\caption{Image classification results on ImageNet-C~\cite{hendrycks2018benchmarking} under imbalanced label distribution shifts scenario. ResNet50 with GN is used for this experiment. We use the accuracy (\%) as the metric. 
$\Delta$Perf. is the performance gap between the original method and another version where our method is incorporated.
% The {best} and \underline{second-best} results are highlighted.
}

\label{tab:label-shift}
% 
% Best results are {highlighted} and second-best results are \underline{underlined}.}
\end{table*}

\begin{table*}[ht]
\centering
\def\arraystretch{1.1}
\setlength{\tabcolsep}{1pt}
{\scriptsize
\begin{tabularx}{\linewidth}{Y|YYYYYCYCYC}
% \hne

\toprule 
Methods & No Adapt & MEMO~\cite{zhang2022memo} & TENT~\cite{wang2021tent} & CoTTA~\cite{wang2022continual} & EATA~\cite{niu2022efficient} & EATA+FATA & SAR~\cite{niu2022towards_SAR} & SAR+FATA & DeYO~\cite{lee2024entropy_deyo} & DeYO+FATA \\
\hline
Accuracy (\%) & 58.35 & 58.15 & 58.36 & 57.57 & 58.58 & 59.71 & 58.37 & 59.18 & 58.56 & 58.64 \\
GMACs & 4.11 & 262.93 & 4.11 & 143.83 & 4.11 & 4.92 & 8.22 & 14.76 & 8.22 & 9.03
\\[-0.5ex]

\bottomrule
\end{tabularx}
}
\caption{Image classification results with computational complexity on Office-Home~\cite{venkateswara2017deep_OfficeHome}. Accuracy is averaged across all domain shift scenarios. ResNet50 with GN is used. 
}
\label{tab:result/office}
\end{table*}
\begin{table}[t]
\centering
\def\arraystretch{1.1}
\setlength{\tabcolsep}{2pt}
\footnotesize {
\begin{tabularx}{\linewidth}{Y|Y|Y}
\toprule
Method & FATA Position & Avg. \\
\hline
No Adapt &  - & 30.62  \\
\hline
DeYO~\cite{lee2024entropy_deyo} & - & 41.67 \\
DeYO+FATA & 0 & 41.52\\
DeYO+FATA & 1 & 42.32\\
DeYO+FATA & 2 & 43.01\\
\rowcolor{lightgray}
DeYO+FATA & 3 & \textbf{45.61}\\[-0.3ex]
\bottomrule
\end{tabularx}
}
\caption{Ablation study on the position of feature augmentation. ResNet50 with GN is used for this experiment.
}
\label{tab:abl/layer}
\end{table}

\vspace{0.25em}
% \subsubsection{Comparison under Batch Size of One}
\noindent\textbf{Comparison under Batch Size of One.}
\cref{tab:bs1} depicts the comparison of the performance in the batch size one scenario. Overall, our proposed method aids the model in adapting more effectively than when the original methods are employed independently. While there is a case where our method exhibits a slight performance drop, the performance improvements generally surpass the amount of drop, indicating that our method remains effective, despite the challenging constraints of the batch size one scenario.

\vspace{0.25em}
% \subsubsection{Comparison under Imbalanced Label Shifts}
\noindent\textbf{Comparison under Imbalanced Label Shifts.}
As shown in \cref{tab:label-shift}, our proposed method demonstrates its effectiveness in the online imbalanced label distribution shift scenario. Remarkably, when our method is combined with EATA, the accuracy improvement reaches an impressive peak of 17.12 points. This substantial enhancement is not limited to EATA alone; our approach also significantly boosts the performance of other methods. 
% By aiding these methods in building more generalizable representations during test-time, our method facilitates considerable performance improvements.

\vspace{0.25em}
\noindent\textbf{Comparison on Office-Home.}
\cref{tab:result/office} shows the comparison of the performance on the Office-Home dataset. FATA consistently enhances the performance with small computational overhead compared to CoTTA, demonstrating the efficiency and robustness of our method.

\begin{table}[t]
\centering
\def\arraystretch{1.05}
\setlength{\tabcolsep}{3pt}
{\footnotesize
\begin{tabularx}{\linewidth}{Y|YY|Y}
% \hne

\toprule
Method & EMA & Std. dev. & Avg. \\
\hline
NP+ & \ding{55} & \ding{55} & 44.84\\
& \checkmark & \ding{55} & 44.96  \\
\rowcolor{lightgray}
FATA (Ours) & \checkmark & \checkmark & \textbf{45.61} \\[-0.3ex]

\bottomrule
\end{tabularx}
}
\caption{Ablation study on the component of feature augmentation. ResNet50 with GN and the
default batch size of 64 are used.
}
\label{tab:abl/faug}
% \vspace{-11pt}
% Best results are {highlighted} and second-best results are \underline{underlined}.}
\end{table}

\subsection{Ablation study}
% \subsubsection{Feature Augmentation}
\noindent\textbf{Location Choice for Feature Augmentation.}
We present the ablation study on the position of feature augmentation in \cref{tab:abl/layer}. Inserting the feature augmentation after the third layer achieves the best performance in average accuracy, therefore, we set \(i=3\) as the default. Embedding it after layer 2 shows comparable performance to that after layer 3. However, for certain corruption types (e.g., Fog), it exhibits a tremendous performance drop, indicating instability. 
Although several works~\cite{fan2023towards_NP, fahes2023poda} have argued that augmenting the features in the shallow layers is the most effective for modifying the style of the input, our results indicate that inserting augmentation after the first layer shows the lowest performance, similar to Li \textit{et al.} (2021)~\cite{li2021simple_SFA}.
% This is because in the TTA setting, those methods could affect the contents of the input, where preserving the shape of the object is more challenging due to the absence of labels and a limited amount of available data, which provide more direct supervision.
% On the contrary, augmenting the features after the third layer would have less impact on the contents of the feature, allowing for better preservation of the underlying information.

\vspace{0.25em}
\noindent\textbf{Component of Feature Augmentation.}
\cref{tab:abl/faug} shows the ablation study on the component of feature augmentation. Compared to NP+~\cite{fan2023towards_NP}, adding EMA and modifying the variance term to the standard deviation improve the accuracy, showing their effectiveness.

\begin{table}[t]
\centering
\def\arraystretch{1.05}
\setlength{\tabcolsep}{15pt}

\footnotesize {
\begin{tabularx}{\linewidth}{cc|Y}
% \hline
\toprule
% \hline
% Time & \multicolumn{15}{l|}{$\; t \xrightarrow{\hspace{1.55\columnwidth}}$} & \multirow{2}{*}{Mean}  \\
% \cline{1-16}
{DeYO} & Augmentation loss & Avg. \\
% \hline
% Time & \multicolumn{15}{l|}{$\; t \xrightarrow{\hspace{1.55\columnwidth}}$} & \multirow{2}{*}{Mean}  \\
% \cline{1-16}
\hline
- & - &  30.62  \\
- & FATA loss (Ours) & 39.35\\
\hline
\checkmark & - & 41.67 \\
%simpleaug
\checkmark & Simple Aug. & 40.30\\
%simplece
%\checkmark & Simple CE & 0.12 & 0.11 & 0.13 & 0.14 & 0.12 & 0.15 & 0.14 & 0.16 & 0.16 & 0.15 & 0.19 & 0.17 & 0.17 & 0.13 & 0.18 & 0.15\\
\checkmark & MSE Loss & 6.54 \\
\checkmark & Simple CE & 44.73 \\
\rowcolor{lightgray}
\checkmark & FATA Loss (Ours) & \textbf{45.61}\\[-0.3ex]

% \rowcolor{lightgray}
% FATA & - \\ [-0.3ex]
% \midrule

\bottomrule
\end{tabularx}
}
\caption{Ablation study on the augmentation loss. ResNet50 with GN is used for this experiment.
}
\label{tab:abl/aug_loss}
% \vspace{-11pt}
% Best results are {highlighted} and second-best results are \underline{underlined}.}
\end{table}

\begin{figure*}[t]
    \centering
    \includegraphics[width=\textwidth]{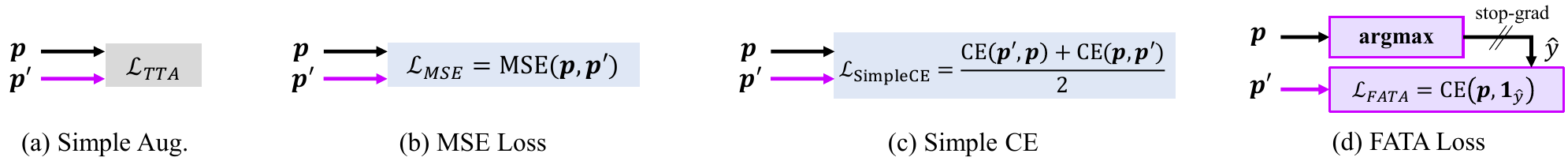}
    \caption{Augmentation losses for the ablation study. $\mathbf{p}$ and $\mathbf{p'}$ denotes $\mathbf{p}_\theta (\mathbf{z})$ and $\mathbf{p}_\theta (\mathbf{z'})$, respectively.}
    \label{fig:abl-losses}
\end{figure*}

% \subsubsection{Augmentation Losses}
\label{subsec:abl-aug_loss}
We ablate the augmentation loss in \cref{tab:abl/aug_loss}. As shown in \cref{fig:abl-losses}a, Simple Augmentation (Simple Aug.) denotes using entropy instead of $\mathcal{L}_\text{FATA}$, i.e., it utilizes entropy loss on the output of augmented feature. MSE Loss denotes using mean square error loss between the outputs, as depicted in \cref{fig:abl-losses}b. Simple CE denotes employing cross-entropy instead of $\mathcal{L}_\text{FATA}$, as shown in \cref{fig:abl-losses}c. In detail, Simple CE excludes the process of obtaining pseudo-labels from our proposed method and uses cross-entropy loss between the output distribution of the augmented features and that of the original feature.

Training a model solely with our method boosts performance by 8.74\% in average accuracy, which is comparable to the performance of the state-of-the-art method, DeYO. Comparing our method with Simple Augmentation, it is evident that augmenting the features rather than the input data is much more effective, as it guides the model to have more generalized representations. % more directly.
% when only a limited amount of samples are available.
In the case of Simple CE, the accuracy on all corruption types are almost zero. This is because naively comparing the output distribution of the original sample and augmented sample leads to the phenomenon of model collapse, as the model learns different features to map to the same output. To avoid this, we do not compare the distribution of the output and adopt pseudo-labeling techniques. 
In the end, incorporating DeYO and our method achieves the best performance.

\begin{figure}[t]
    \centering
    \begin{subfigure}{0.49\linewidth}
        \includegraphics[width=\linewidth]{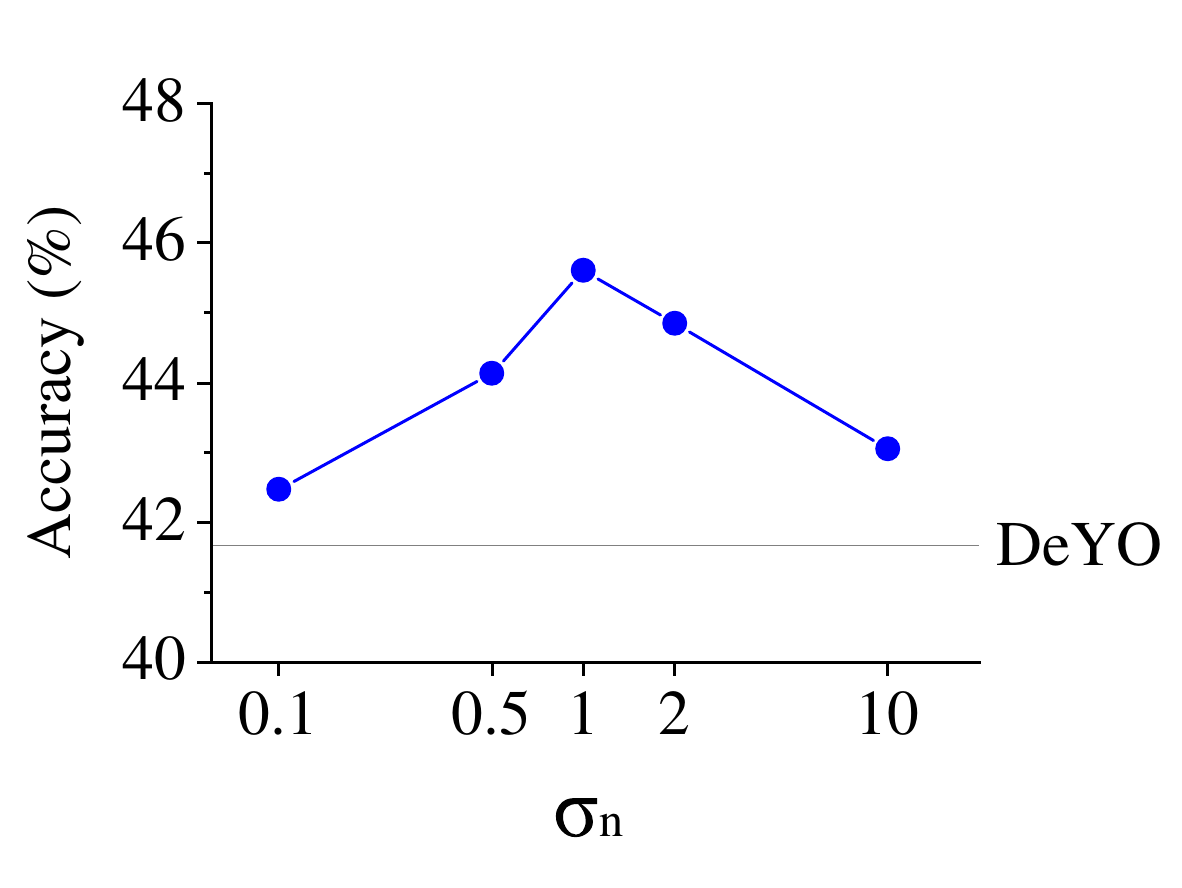}
        \caption{Parameter sensitivity of $\sigma_n$.}
        \label{fig:result-param-sigma}
    \end{subfigure}
    \hfill
    \begin{subfigure}{0.49\linewidth}
        \includegraphics[width=\linewidth]{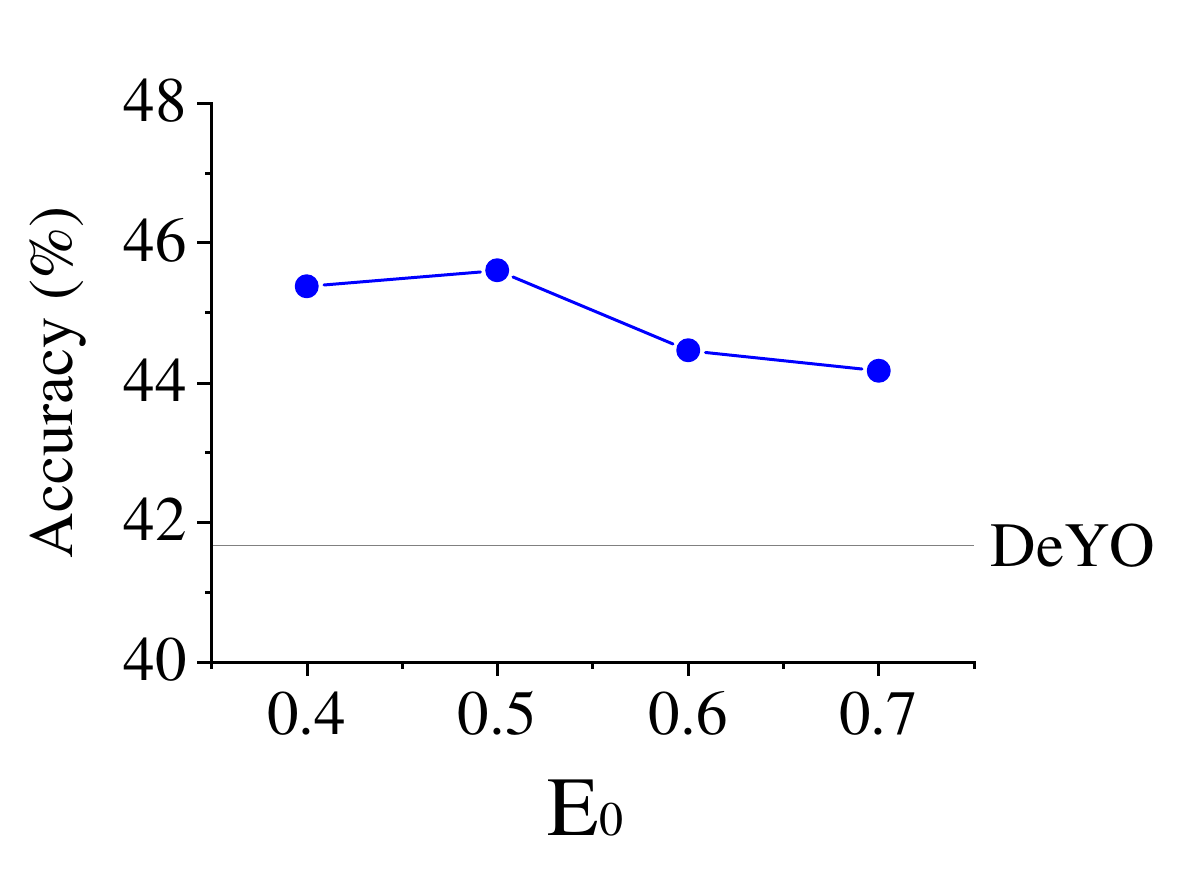}
        \caption{Parameter sensitivity of $E_0$.}
        \label{fig:result-param-E0}
    \end{subfigure}
    \caption{Hyperparameter Sensitivity.}
    \label{fig:result-param}
\end{figure}

\vspace{0.25em}
% \subsubsection{Hyperparameter Sensitivity}
\noindent\textbf{Hyperparameter Sensitivity.}
We demonstrate the sensitivity of the hyperparameters in \cref{fig:result-param}. 
We conduct experiments to assess the sensitivity of two hyperparameters: the entropy threshold $E_0$ and the standard deviation for the noise $\sigma_n$. As shown in \cref{fig:result-param-sigma}, the highest accuracy is achieved with  $\sigma_n=1$, although other values, ranging from $\sigma_n=0.1$ to $10$, consistently outperform DeYO. Similarly, \cref{fig:result-param-E0} shows that an entropy threshold of $E_0=0.5$ yields the best results, with other values also surpassing the performance of DeYO. These findings show the robustness of our methods across a variety of hyperparameter settings.

% \newpage
\subsection{Discussion}

\begin{figure}[t]
    \centering
    \includegraphics[width=.95\linewidth]{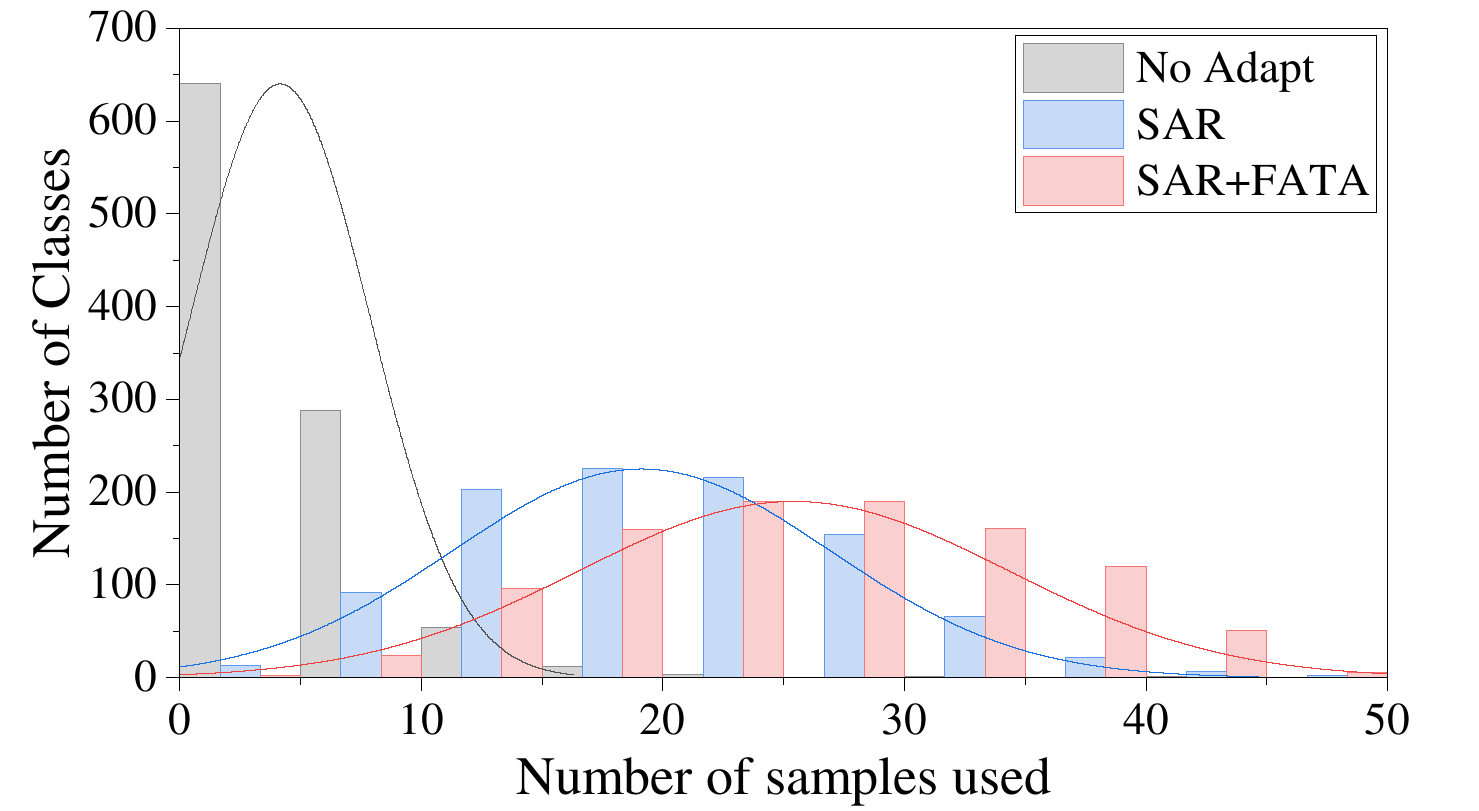}
    \caption{Count of classes for each number group of selected samples. 
    %No Adapt model, SAR~\cite{niu2022towards_SAR} and SAR+FATA are compared.
    }
    \label{fig:result-classes}
\end{figure}

% \subsubsection{The Number of Samples Used} 
\noindent\textbf{The Number of Samples Used.} 
\cref{fig:result-classes} compares the count of classes for each group of selected samples for the No Adapt, SAR, and SAR+FATA. With SAR, a significantly larger number of classes are selected more than 5 times, whereas without any TTA method, many classes are sampled 5 times or fewer. When our method is incorporated with SAR, considerably more classes are sampled more frequently compared to SAR alone. This increased sampling frequency contributes to the performance boost provided by FATA, as shown by \cref{fig:intro/corr}, which depicts the positive correlation between the sampling frequency and the accuracy. 

% demonstrates that FATA increases the number of used samples.

\vspace{0.25em}
% \subsubsection{Analysis on FATA Loss}
\noindent\textbf{Analysis on FATA Loss.}
The FATA loss demonstrates its effectiveness with the experimental results in Sec.~\ref{subsec:abl-aug_loss}, in contrast to MSE Loss and Simple CE. With MSE Loss, the model tends to collapse in order to align two distribution, as it is the trivial solution to achieve alignment under the random feature augmentation by producing the same output for any input. 
Simple CE improves the model performance, but less than the FATA loss, although the main difference is the hard label generated by the argmax operator.
In contrast, the FATA loss mitigates the trivial solution by replacing the output from the original feature with a pseudo-label. Further research should further develop a theoretical analysis for the FATA loss.

% \noindent\textbf{Further Analysis on FATA Loss. }
% Despite of its effectiveness, there lacks a theoretical analysis on FATA loss 

\vspace{0.25em}
% \subsubsection{Limitation} 
\noindent\textbf{Limitation.}
% collapse가 나는 경우는 우리 방법론을 사용하여도 100프로 막지는 못한다. 할 떄도 있다. further exploration이 필요하다.
% \noindent\textbf{Collapse Prevention. }
Although our method effectively enhances the model performance, our method does not have the capability to prevent the collapse phenomenon. For example, as shown in \cref{tab:label-shift}, the accuracy of SAR on fog corruption type is 2.77\%, while it decreases to 1.56\% with SAR+FATA. Future research should explore a method to prevent the collapse phenomenon.

\section{Conclusion}
In this paper, we propose a test-time adaptation method named Feature Augmented Test-Time Adaptation (FATA). This method fully utilizes the target samples through the feature augmentation technique, addressing the issue of limited samples from sample selection based entropy minimization methods. FATA can be seamlessly integrated into any method that employing entropy-based sampling, allowing a model to leverage target samples more effectively with reliably selected samples. FATA boosts the performance of existing methods across various network architectures. Extensive experiments, including challenging scenarios, validate the effectiveness and robustness of FATA.

%%%%%%%%% REFERENCES
{\small
\bibliographystyle{ieee_fullname}
\bibliography{main}
}

\end{document}

\newcolumntype{Y}{>{\centering\arraybackslash}X}
\newcolumntype{Z}{>{\hsize=.75\hsize\linewidth=\hsize\centering\arraybackslash}X}
\newcolumntype{V}{>{\hsize=1.25\hsize\linewidth=\hsize\centering\arraybackslash}X}
%%%%%%%%% TITLE - PLEASE UPDATE
\title{Feature Augmentation based Test-Time Adaptation\\
Supplementary Material}

\author{Younggeol Cho\footnotemark[1] \qquad
Youngrae Kim\footnotemark[1] \qquad
Junho Yoon \qquad
Seunghoon Hong \qquad
Dongman Lee\\
KAIST\\
{\tt\small \{rangewing, youngrae.kim, , seunghoon.hong, dlee\}@kaist.ac.kr} 
% For a paper whose authors are all at the same institution,
% omit the following lines up until the closing ``}''.
% Additional authors and addresses can be added with ``\and'',
% just like the second author.
% To save space, use either the email address or home page, not both
}

\maketitle
\setcounter{section}{0}
\setcounter{table}{0}
\setcounter{figure}{0}
\setcounter{equation}{0}
\renewcommand{\thesection}{\Alph{section}}
\renewcommand{\thetable}{\Alph{table}}
\renewcommand{\thefigure}{\Alph{figure}}
\renewcommand{\theenumi}{\;\;\Alph{enumi}}
\renewcommand{\theequation}{\Alph{equation}}

\begin{table*}[bh!]
\centering
\def\arraystretch{1.2}
\setlength{\tabcolsep}{2pt}
\footnotesize {
\begin{tabularx}{\textwidth}{c|*{12}{Y}|Yc}
% \hline
\toprule
% \hline
% Time & \multicolumn{15}{l|}{$\; t \xrightarrow{\hspace{1.55\columnwidth}}$} & \multirow{2}{*}{Mean}  \\
% \cline{1-16} 
Method & Ar→Cl & Ar→Pr & Ar→Re & Cl→Ar & Cl→Pr & Cl→Re & Pr→Ar & Pr→Cl & Pr→Re & Re→Ar & Re→Cl & Re→Pr & Average & $\Delta$Perf. \\
\hline
No Adapt & 41.31 & 64.74 & 74.75 & 51.71 & 60.94 & 63.58 & 51.3 & 37.21 & 73.63 & 64.94 & 40.25 & 75.85 & 58.35 & \\
MEMO~\cite{zhang2022memo} & 42.18 & 65.65 & 75.03 & 49.94 & 60.26 & 62.82 & 50.31 & 37.25 & 72.44 & 64.81 & 40.37 & 76.73 & 58.15 & \\
TENT~\cite{wang2021tent} & 41.4 & 64.74 & 74.75 & 51.71 & 60.89 & 63.64 & 51.3 & 37.16 & 73.65 & 65.02 & 40.21 & 75.87 & 58.36 & \\
CoTTA~\cite{wang2022continual} & 41.28 & 64.79 & 71.86 & 51.71 & 60.91 & 63.6 & 51.3 & 37.21 & 72.11 & 64.94 & 40.3 & 70.87 & 57.57 & \\
EATA~\cite{niu2022efficient} & 41.99 & 64.83 & 74.87 & 51.71 & 60.96 & 63.64 & 51.42 & 37.82 & 73.67 & 65.1 & 41.03 & 75.94 & 58.58 & \\ 
\rowcolor{lightgray}
EATA+FATA  & 42.2 & 65.74 & 74.39 & 53.61 & 61.66 & 65.34 & 53.61 & 39.11 & 74.27 & 67.08 & 43.09 & 76.41 & 59.71 & +1.13 \\
SAR~\cite{niu2022towards_SAR} & 41.37 & 64.77 & 74.78 & 51.67 & 60.91 & 63.64 & 51.34 & 37.14 & 73.63 & 65.02 & 40.25 & 75.87 & 58.37 &  \\
\rowcolor{lightgray}
SAR+FATA  & 42.02 & 64.63 & 74.73 & 53.32 & 61.59 & 64.82 & 51.63 & 38.72 & 73.88 & 65.51 & 42.47 & 76.82 & 59.18 & +0.81 \\
DeYO~\cite{lee2024entropy_deyo}  & 41.63 & 64.7 & 74.71 & 51.96 & 61.07 & 63.87 & 51.34 & 37.71 & 73.74 & 65.1 & 40.62 & 76.3 & 58.56 &  \\
\rowcolor{lightgray}
DeYO+FATA & 41.74 & 64.97 & 74.73 & 51.96 & 61.34 & 63.99 & 51.55 & 37.73 & 73.74 & 64.94 & 40.82 & 76.19 & 58.64 & +0.08
  \\

\bottomrule
\end{tabularx}
}
\caption{Image classification results on Office-Home~\cite{venkateswara2017deep_OfficeHome}. ResNet50 with GN is used for this experiment. We use the accuracy (\%) as the metric. 
$\Delta$Perf. is the performance gap between the original method and another version where our method is incorporated.
% The {best} and \underline{second-best} results are highlighted.
}

\label{tab:supp/office}
% 
% Best results are {highlighted} and second-best results are \underline{underlined}.}
\end{table*}

\begin{table*}[b]
\centering
\def\arraystretch{1.3}
\setlength{\tabcolsep}{2pt}
\scriptsize {
\begin{tabularx}{\textwidth}{c|*{3}{Y}|*{4}{Y}|*{4}{Y}|*{4}{Y}|c}
\toprule
\multirow{2}{*}{Method} & 
\multicolumn{3}{c|}{Noise} & \multicolumn{4}{c|}{Blur} & \multicolumn{4}{c|}{Weather} & \multicolumn{4}{c|}{Digital} & \multirow{2}{*}{ Avg.}
\\[-0.7ex]
& {Gauss.} &  {Shot} &  {Impul.} &  {Defoc.} &  {Glass} &  {Motion} &  {Zoom} &  {Snow} &  {Frost} &  {Fog} &  {Brit.} &  {Contr.} &  {Elastic} &  {Pixel} &  {JPEG}  &   \\
\hline
No Adapt &  17.98 & 19.84 & 17.88 & 19.75 & 11.35 & 21.42 & 24.92 & 40.43 & 47.30 & 33.59 & 69.28 & 36.27 & 18.61 & 28.40 & 52.28 & 30.62  \\[-0.3ex]
\hline
DeYO~\cite{lee2024entropy_deyo} & 39.46 & 41.90 & 41.03 & {22.27} & {24.11} & 38.48 & 37.87 & 50.51 & 49.59 & 1.43 & 73.17 & 49.95 & 41.54 & 55.96 & 57.82 & 41.67 \\ [-0.3ex]
DeYO+FATA (Layer 0) & 35.58 & 38.56 & 36.51 & 20.32 & 17.61 & {41.07} & 40.53 & 48.89 & 48.40 & {54.22} & 71.25 & 49.74 & 8.08 & 55.29 & 56.76 & 41.52\\
DeYO+FATA (Layer 1) & {39.84} & 42.11 & 40.47 & 21.84 & 23.68 & {39.93} & {41.17} & {55.03} & {50.82} & 1.09 & 72.86 & {51.13} & 39.57 & {57.26} & {57.97} & 42.32\\
DeYO+FATA (Layer 2) & {39.72} & {42.77} & {41.03} & 21.44 & 23.04 & 39.74 & {42.76} & {54.51} & {52.04} & 0.84 & {73.57} & {51.67} & {45.92} & {57.89} & {58.16} & {43.01}\\
\rowcolor{lightgray}
DeYO+FATA (Layer 3) & 39.71 & {42.49} & {41.37} & {22.29} & {24.46} & 38.90 & 38.31 & 51.23 & 50.02 & {56.31} & {73.19} & 50.03 & {42.10} & 55.99 & 57.79 & {45.61}\\[-0.3ex]

\bottomrule
\end{tabularx}
}
\caption{Ablation study on the position of feature augmentation. ResNet50 with GN is used for this experiment.
}
\label{tab:supp-abl/layer}
\end{table*}

\begin{table*}[t]
\centering
\def\arraystretch{1.3}
\setlength{\tabcolsep}{1.0pt}

\scriptsize {
\begin{tabularx}{\textwidth}{cc|*{3}{Y}|*{4}{Y}|*{4}{Y}|*{4}{Y}|c}
% \hline
\toprule
% \hline
% Time & \multicolumn{15}{l|}{$\; t \xrightarrow{\hspace{1.55\columnwidth}}$} & \multirow{2}{*}{Mean}  \\
% \cline{1-16}
\multirow{2}{*}{DeYO} & \multirow{2}{*}{ Augmentation loss} &
\multicolumn{3}{c|}{Noise} & \multicolumn{4}{c|}{Blur} & \multicolumn{4}{c|}{Weather} & \multicolumn{4}{c|}{Digital} & \multirow{2}{*}{Avg.}
\\[-0.7ex]
% \hline
% Time & \multicolumn{15}{l|}{$\; t \xrightarrow{\hspace{1.55\columnwidth}}$} & \multirow{2}{*}{Mean}  \\
% \cline{1-16}
 & & {Gauss.} &  {Shot} &  {Impul.} &  {Defoc.} &  {Glass} &  {Motion} &  {Zoom} &  {Snow} &  {Frost} &  {Fog} &  {Brit.} &  {Contr.} &  {Elastic} &  {Pixel} &  {JPEG}  & \\
\hline
- & - &  17.98 & 19.84 & 17.88 & 19.75 & 11.35 & 21.42 & 24.92 & 40.43 & 47.30 & \underline{33.59} & 69.28 & 36.27 & 18.61 & 28.40 & 52.28 & 30.62  \\[-0.3ex]
\hline
- & FATA loss (Ours) & 37.92 & 39.99 & 38.70 & \textbf{25.96} & 22.21 & 36.06 & 37.77 & \textbf{51.99} & \textbf{51.33} & 2.85 & 73.11 & 48.83 & 13.10 & 53.78 & 56.73 & 39.35\\
\hline
\checkmark & - & \underline{39.46} & \underline{41.90} & \underline{41.03} & {22.27} & \underline{24.11} & \underline{38.48} & \underline{37.87} & 50.51 & 49.59 & 1.43 & \underline{73.17} & \underline{49.95} & \underline{41.54} & \underline{55.96} & \textbf{57.82} & 41.67 \\ [-0.3ex]
%simpleaug
\checkmark & Simple Aug. & 38.19 & 41.31 & 39.31 & 21.64 & 23.43 & 36.96 & 34.93 & 48.72 & 46.50 & 1.14 & 72.90 & 49.19 & 37.43 & 55.39 & 57.39 & 40.30\\
%simplece
%\checkmark & Simple CE & 0.12 & 0.11 & 0.13 & 0.14 & 0.12 & 0.15 & 0.14 & 0.16 & 0.16 & 0.15 & 0.19 & 0.17 & 0.17 & 0.13 & 0.18 & 0.15\\
\checkmark & MSE loss & 0.22 & 0.46 & 0.57 & 0.88 & 0.89 & 0.57 & 2.15 & 2.00 & 7.49 & 8.18 & 61.78 & 6.37 & 0.83 & 2.19 & 3.52 & 6.54 \\
\rowcolor{lightgray}
\checkmark & FATA Loss (Ours) & \textbf{39.71} & \textbf{42.49} & \textbf{41.37} & \underline{22.29} & \textbf{24.46} & \textbf{38.90} & \textbf{38.31} & \underline{51.23} & \underline{50.02} & \textbf{56.31} & \textbf{73.19} & \textbf{50.03} & \textbf{42.10} & \textbf{55.99} & \underline{57.79} & \textbf{45.61}\\[-0.3ex]

% \rowcolor{lightgray}
% FATA & - \\ [-0.3ex]
% \midrule

\bottomrule
\end{tabularx}
}
\caption{Ablation study on the augmentation loss. ResNet50 with GN is used for this experiment.
}
\label{tab:supp-abl/aug_loss}
% \vspace{-11pt}
% Best results are {highlighted} and second-best results are \underline{underlined}.}
\end{table*}

% \section{Complete Results}
Due to the limited space of the main paper, we provide more experimental results and implementation details in the supplementary material.

\begin{figure}[t]
    \centering
    \includegraphics[width=\linewidth]{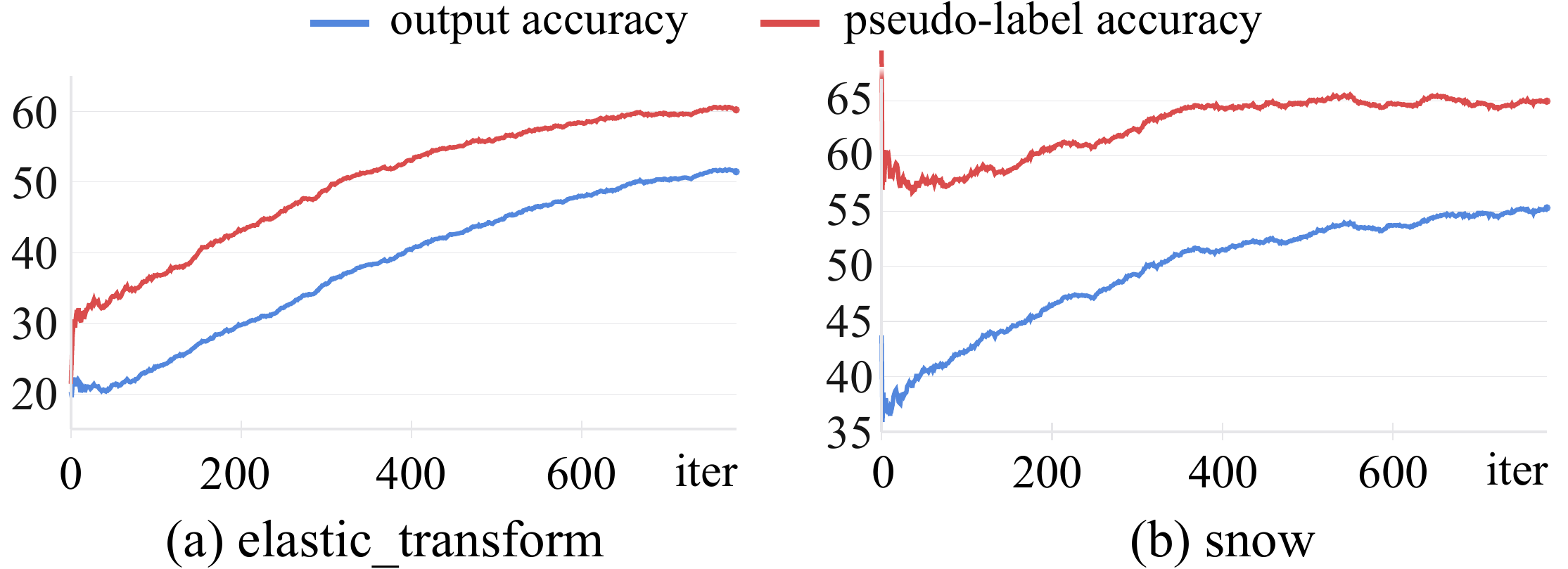}
    
    \caption{Pseudo-label accuracy (\%) on ImageNet-C. We use DeYO+FATA with ResNet50-GN for this experiment. Exponential moving average ($p=0.99$) is applied.
    \vspace{1em}}
    \label{fig:rebuttal-pseudolabels}
\end{figure}

\vspace{0.5em}
\noindent\textbf{Accuracy of pseudo-labels.} 
\cref{fig:rebuttal-pseudolabels} shows the accuracy of pseudo-labels as the model adapts. The pseudo-label accuracy is high compared to the output accuracy and improves over time.

\vspace{0.5em}
\noindent\textbf{Further experimental results.} 
We present the full results of Tab. 4, Tab. 5, and Tab. 7 in the main paper, as shown in \cref{tab:supp/office}, \cref{tab:supp-abl/layer}, and \cref{tab:supp-abl/aug_loss}, respectively.

\vspace{0.5em}
\noindent\textbf{Implementation details.} 
We provide the full data of the hyperparameters in Tab.~\ref{tab:hyperparams}.
% Sec.~\ref{sec:exp}.

% \input{tab/supp-abl-layer}
% \input{tab/abl-loss.old}
% \subsection{Complete Results}
% \input{tab/supp-abl-aug} % CANNOT USE

% Ablation study on the position of feature augmentation

% Ablation study on the component of feature augmentation

% \section{Further Implementation Detail}

\begin{table*}[h!]
\centering
\def\arraystretch{1.1}
{\small
\begin{tabularx}{\linewidth}{cYY}
\hline
\toprule
Parameter & {Meaning} & Value \\
\hline
$i$ & {Layer to inject the feature augmentation} & 3 (ResNet50), 11 (ViT-B) \\
$E_0$ & {Entropy threshold for sample filtering} & $0.5\ln{C}$ \\
$E_\omega$ & {Normalization factor for sample weighting} & $0.4\ln{C}$ \\
$\sigma_n$ & {Standard deviation for the noise} & 1.0 \\
$\lambda_{EMA}$ & {Smoothing factor for exponential moving average} & 0.95 \\
$B$ & {Default batch size} & 64 \\
$\eta$ &
{Learning rate} & 0.0005 (ResNet50, $B=64$)  \\
& & 0.001 (ViT-B, $B=64$)\\
&& 0.00025 (ResNet50, $B=1$) \\
& & 0.000016 (ViT-B, $B=1$)\\
% $$
\bottomrule
\end{tabularx}
}
\caption{Hyperparameters for the experiment.}
% \vspace{-11pt}
% Best results are \textbf{highlighted} and second-best results are \underline{underlined}.}
\label{tab:hyperparams}
\end{table*}

\vspace{0.5em}
\noindent\textbf{Rationale for replacing variance with standard deviation in Eq. (3).}
Eq. (3) is derived from Normalization Perturbation (NP)~\cite{fan2023towards_NP}, which is formulated as follows:

\begin{equation}
\mathbf{z'} = \alpha \sigma_c \cdot \frac{\mathbf{z}-\mu_c}{\sigma_c} + \beta\mu_c = \alpha \mathbf{z} + (\beta-\alpha)\mu_c ,
\label{eq:supp/NP}
\end{equation}
where $\alpha$, $\beta \in \mathbb{R}^{B\times C}$ are the random noises sampled from $N(\mathbf{I}, \sigma_{n}\mathbf{I})$, $\mu_c, \sigma_c$ are the channel-wise feature mean and standard deviation, respectively. As formulated by Eq. (3), NP+ introduces the term $\delta = \text{Var}(\mu_c) / \text{max}(\text{Var}(\mu_c))$ to control the magnitude of the noises $\alpha$ and $\beta$. We replace variance with standard deviation because the noises are related to standard deviation $\sigma_c$, not variance, as shown in \cref{eq:supp/NP}.

%%%%%%%%% REFERENCES
{\small
\bibliographystyle{ieee_fullname}
\bibliography{main}
}